\documentclass{article}

\PassOptionsToPackage{numbers, compress}{natbib}

\usepackage[preprint]{neurips_2026}

\workshoptitle{AI for Meta-Science workshop}

\usepackage[utf8]{inputenc} % allow utf-8 input
\usepackage[T1]{fontenc}    % use 8-bit T1 fonts
\usepackage{hyperref}       % hyperlinks
\usepackage{url}            % simple URL typesetting
\usepackage{booktabs}       % professional-quality tables
\usepackage{amsfonts}       % blackboard math symbols
\usepackage{nicefrac}       % compact symbols for 1/2, etc.
\usepackage{microtype}      % microtypography
\usepackage{xcolor}         % colors

\usepackage{graphicx}
\usepackage{amsmath}
\usepackage{xspace}
\usepackage{enumitem}
\usepackage{subcaption}
\usepackage[hide]{todo}
\usepackage{placeins}

\usepackage{listings}
\newcommand{\qt}[1]{``#1''}
\newcommand{\ATD}{ATD\xspace} %Artificial Text Detection
\usepackage{multirow}       % required by the tables imported from the peer-review paper
\newcommand{\method}{Self-Conditioning\xspace}
\newcommand{\scoringmodel}{Llama-3.1-8B-IT\xspace}

\title{
How Much Were You Told? Measuring External Information in Peer Reviews
}

\author{Matthieu Dubois \\
  Sorbonne Université, CNRS, ISIR \\
  \texttt{duboism\textit{[at]}isir.upmc.fr} \\\AND
  Pablo Piantanida \\
  International Laboratory on Learning Systems (ILLS) \\
  Quebec AI Institute (MILA) \\
  CNRS, CentraleSupélec, Université Paris-Saclay \\
  \texttt{pablo.piantanida[\textit{at}]mila.quebec} \\\And
  François Yvon \\
  Sorbonne Université, CNRS, ISIR \\
  \texttt{yvon[\textit{at}]isir.upmc.fr} \\
}

\begin{document}

%\raggedbottom % COMMENT OUT FOR SUBMISSION
\maketitle

%% ############################################################################
%% ############################################################################
%% ###                                                                      ###
%% ###   REWRITTEN BLOCK -- ABSTRACT
%% ###   Reframed on "external information"; essays and reference letters dropped.
%% ###                                                                      ###
%% ############################################################################
%% ############################################################################

\begin{abstract}
Conference policies distinguish using Large Language Models (LLMs) to polish
one's own review from delegating the critique, but current Artificial Text
Detection (\ATD) methods largely measure surface form rather than the origin of
its content. We instead measure the \textbf{external information} carried by a
review: information not explained by the reviewed paper and a generic reviewing
instruction. We propose \method, an unsupervised information-theoretic estimator
that compares the likelihood of a review under its production context with its
likelihood when that context is augmented with hints extracted from the review
itself. On the IntelLabs peer-review benchmark, \method separates fully-delegated
from machine-polished reviews with AUC up to $1.0$ while remaining largely
insensitive to surface rewriting. Moreover, as generators receive increasing
amounts of externally-provided information, their scores move monotonically
towards the human regime, unlike standard \ATD baselines. High-temperature
sampling can evade the estimator, but at the cost of output quality.
\end{abstract}

%% ############################ END REWRITTEN BLOCK ###########################

%% ############################################################################
%% ############################################################################
%% ###                                                                      ###
%% ###   REWRITTEN BLOCK -- SECTION 1 -- INTRODUCTION
%% ###   Merged the two redundant openings (policy stakes + "binary distinction"), reframed on external information, dropped essays/reference letters, new contribution list.
%% ###                                                                      ###
%% ############################################################################
%% ############################################################################

\section{Introduction} \label{sec:intro}

Peer reviewing is a cornerstone of scientific publications, but growing submission
volumes, particularly in AI \citep{sculley-etal-2018-avoiding}, make reviewing
increasingly time-consuming and may encourage overburdened reviewers to delegate
it to LLMs. This is especially relevant as LLM-generated reviews continue to
improve \citep{yu-etal-2024-automated, darcy-etal-2024-MARGMultiAgentReview,
zhu-etal-2025-deepreview, thakkar-etal-2026-largescale},
despite reviewers being expected to perform the substantive work themselves
\citep{jefferson-etal-2002-peer, rogers-augenstein-2020-improve,
church-etal-2025-peer}.

Major Machine Learning and Natural Language Processing venues therefore
distinguish \emph{writing assistance} from delegation. ICLR, NeurIPS, and *ACL
generally prohibit sharing anonymous submissions with models while permitting
assistance with grammar or clarity,\footnote{See, for instance, the ACL Policy
on Publication Ethics
(\url{https://www.aclweb.org/adminwiki/index.php/ACL_Policy_on_Publication_Ethics}).}
whereas ICML 2026 permits privacy-compliant models to aid
comprehension,\footnote{\url{https://icml.cc/Conferences/2026/LLM-Policy}}
but not to generate criticisms. Despite these differences, the boundary is
essentially the same: \emph{the wording may be machine-made, the underlying
content may not}. Enforcing this distinction remains difficult despite
countermeasures \citep{rao-etal-2025-detecting} and dedicated detection methods
\citep{kumar-etal-2024-quis}.

Existing Artificial Text Detection (\ATD) addresses a different question
\citep{wu-etal-2025-survey}. Consider two reviewers: one identifies a paper's
strengths and weaknesses and asks an LLM only to polish the resulting review;
the other gives the paper to an LLM and asks it to produce the critique.
Current detectors, including methods designed for hybrid text
\citep{thai-etal-2026-editlens, wang-etal-2025-real}, may flag both as
AI-generated although only the latter delegates the substantive reviewing work.
\ATD asks whether a model shaped the \emph{surface}; policy compliance depends
on who or what supplied the \emph{content}.

We therefore ask a different question: \emph{how much would a model need to be
told, beyond the manuscript and generic reviewing instructions, to produce this
review?} For a fully-delegated review, the generator receives no additional
information; a human-written or machine-polished review instead contains
information supplied beyond that context. We call this quantity the
\textbf{external information} of a review and estimate it using an
information-theoretic criterion computed by a proxy language model. The score
is source-agnostic: it measures what the production context does not explain,
not whether the additional information came from a human, retrieval, or another
system.

\paragraph{Contributions.}
We (1) formulate peer-review assessment as estimating external information
beyond the paper and generic reviewing instructions; (2) propose \method, an
unsupervised estimator that separates fully-delegated from machine-polished
reviews; and (3) show that its score varies monotonically with controlled
amounts of externally-provided information, while characterising important
edge cases. We release our code and data at
\url{https://github.com/BaggerOfWords/Measuring-External-Information}.

%% ############################ END REWRITTEN BLOCK ###########################

%% ############################################################################
%% ############################################################################
%% ###                                                                      ###
%% ###   REWRITTEN BLOCK -- SECTION 2 -- RELATED WORK
%% ###   Seven paragraphs compressed to four; benchmarks paragraph trimmed; positioning rewritten around external information.
%% ###                                                                      ###
%% ############################################################################
%% ############################################################################

\section{Related Work} \label{sec:related_work}

\paragraph{Peer reviewing, NLP, and AI.}
PeerRead \citep{kang-etal-2018-dataset} and NLPeer
\citep{dycke-etal-2023-nlpeer} provide corpora for computational studies of
peer review, while recent work studies unconstructive reviewing
\citep{purkayastha-etal-2025-lazyreview} and review-derived inquiries
\citep{baumgartner-etal-2025-peerqa}. LLMs are increasingly used to generate
reviews \citep{darcy-etal-2024-MARGMultiAgentReview,
zhu-etal-2025-DeepReviewImprovingLLMbased,
mann-etal-2026-AIFutureAcademic}, although they struggle with methodological
novelty \citep{sahu-etal-2025-ReviewerTooShouldAI, shin-etal-2025-mind} and
replicate human biases \citep{jin-etal-2024-agentreview}. Undisclosed use has
consequently motivated countermeasures \citep{rao-etal-2025-detecting} and
detectors for fully-generated \citep{yu-etal-2026-is-your-paper} and mixed
reviews \citep{kumar-etal-2025-mixrevdetect, chen-etal-2026-coconuts}.

\paragraph{AI-generated text detection.}
Most \ATD work distinguishes human from machine-generated text
\citep{gehrmann-etal-2019-gltr}, with extensions to hybrid writing
\citep{abassy-etal-2024-llm, kumar-etal-2025-mixrevdetect}, generator-specific
detection \citep{liu2024detectability}, and source attribution
\citep{liOriginTracingDetecting2023}. Supervised detectors perform well
in-domain \citep{guo-etal-2023-HowCloseChatGPT,
verma-etal-2024-ghostbuster} but degrade under shifts in generator, prompt,
domain \citep{antoun-etal-2024-text, kuznetsov-etal-2024-robust}, or decoding
\citep{dubois-etal-2025-sampling}. Zero-shot alternatives exploit curvature
\citep{mitchell-etal-2023-DetectGPT, bao-etal-2024-fastdetectgpt},
contrastive likelihoods \citep{hans-etal-2024-SpottingLLMsBinoculars}, or
multiple observers \citep{dubois-etal-2025-mosaic-multiple}.

\paragraph{Editing and adversarial evaluation.}
Hybrid-text methods use rewriting discrepancies
\citep{yang-etal-2024-dnagpt, mao-etal-2024-raidar} or explicitly estimate AI
editing, as in EditLens \citep{thai-etal-2026-editlens}. Benchmarks including
MAGE and RAID \citep{li-etal-2024-mage, dugan-etal-2024-raid,
wang-etal-2024-stumbling} expose detector sensitivity to adversarial prompting
\citep{kumarage-etal-2023-reliable, koike-etal-2024-prompt,
lu-etal-2024-guided} and paraphrasing \citep{krishna-etal-2023-paraphrasing},
under which detection remains unreliable
\citep{sadasivan-2024-AItext, pudasaini-etal-2025-benchmarking,
voight-kampff-2025}. \citet{xie-etal-2026-measuring} instead study mutual
information between generated text and its known prompt.

\paragraph{Positioning.}
Our formulation draws on the coding interpretation of language-model
likelihoods \citep{deletang-etal-2024-language} and information-theoretic prompt
selection \citep{sorensen-etal-2022-information}. Unlike provenance, source
attribution, or AI-editing measures, we estimate the
\emph{externally-provided information} embodied in a review relative to its
production context. \method is therefore a graded, context-relative measure
rather than a conventional detector.

%% ############################ END REWRITTEN BLOCK ###########################

%% ############################################################################
%% ############################################################################
%% ###                                                                      ###
%% ###   REWRITTEN BLOCK -- SECTION 3 -- PROBLEM FORMULATION / METHOD
%% ###   Reframed on external information; PI regime and the FD -> PI -> MP ~ H continuum added; MDL discussion tightened; hint-family list reduced to two sentences; token-level figure moved to Appendix.
%% ###                                                                      ###
%% ############################################################################
%% ############################################################################

\section{Measuring External Information} \label{sec:problem}

\subsection{Review-production regimes}
\label{sec:regimes}

We distinguish four regimes. \emph{Human-authored} reviews (\textit{H}) are
OpenReview reviews submitted before 2022. \emph{Machine-polished} reviews
(\textit{MP}) are human reviews edited by an LLM for grammar or clarity.
\emph{Partially-informed} reviews (\textit{PI}) are generated from the paper
plus $k$ bullet points extracted from a human review along with its length, while
\emph{fully-delegated} reviews (\textit{FD}) are generated from the paper, conference guidelines and the final decision.

They form a continuum in externally supplied information:
\[
\textit{FD} \;\longrightarrow\; \textit{PI}_{k=1}
\;\longrightarrow\; \cdots \;\longrightarrow\;
\textit{PI}_{k=8} \;\longrightarrow\; \textit{MP} \approx \textit{H}.
\]
A valid measure should recover this ordering. Section~\ref{sec:results}
evaluates its endpoints and Section~\ref{ssec:graded_effort} the intermediate
\textit{PI} regimes. As no standard benchmark covers this new formulation, we
use controlled settings as a first approximation.

\subsection{External information as context-relative information gain}
\label{sec:external-information}

Let $C$ denote the \emph{baseline production context} and
$T=(Y_1,\ldots,Y_n)$ the observed text. For peer review, $C$ contains the paper
and generic reviewing instructions. Production may additionally depend on
information unavailable in $C$ (e.g., judgments, critiques, or
external knowledge) which we represent by a latent variable $U$:
\begin{equation}
    T \sim P_{T\mid C,U}.
    \label{eq:latent-production}
\end{equation}
Here $U$ denotes information supplied beyond $C$, irrespective of provenance:
it may come from human reasoning, retrieval, tools, or other sources. Our goal
is not to reconstruct $U$ or identify the author, but to quantify information
in $T$ not explained by $C$.

\paragraph{One-shot information gain.}
For an individual realization $(t,c)$, suppose additional information $x$ is
revealed. Its explanatory value is the conditional information density
\begin{equation}
    \imath(t;x\mid c)
    :=
    \log
    \frac{P(t\mid c,x)}
         {P(t\mid c)}.
    \label{eq:conditional-information-density}
\end{equation}
This measures how much $x$ increases the likelihood of $t$; equivalently, it is
the reduction in ideal description length. Appendix~\ref{app:theory} gives the
full coding interpretation.

\paragraph{Controlled self-disclosure.}
Because $U$ is unobserved, we cannot evaluate $\imath(t;u\mid c)$ directly.
Instead, let $X\sim q_\phi(\cdot\mid T)$ be a randomised
\emph{disclosure channel} revealing a fixed-budget fragment of $T$ (e.g.,
words, $n$-grams, keywords, or high-surprisal spans). Since $X$ is constructed
only from $T$, $(C,U)\rightarrow T\rightarrow X$.

We define the ideal \emph{Self-Conditioning information gain} as
\begin{equation}
    \mathcal S_\phi(t\mid c)
    :=
    \frac{1}{|t|}
    {\mathbb{E}}_{X\sim q_\phi(\cdot\mid t)}
    \left[
        \log
        \frac{P(t\mid c,X)}
             {P(t\mid c)}
    \right].
    \label{eq:ideal-sc}
\end{equation}
Normalization by $|t|$ permits comparisons across lengths.
$\mathcal S_\phi(t\mid c)$ measures how much a fixed disclosure improves the
explanation of $t$ beyond $c$ and is therefore explicitly context-relative.

\paragraph{Relation to external information.}
The corresponding unnormalised population gain is $I(T;X\mid C)$
(Appendix~\ref{app:theory}). Under
$(C,U)\rightarrow T\rightarrow X$,
\begin{equation}
    \underbrace{I(T;X\mid C)}_{\text{Self-Conditioning information}}
    =
    \underbrace{I(U;X\mid C)}_{\text{recoverable external contribution}}
    +
    \underbrace{I(T;X\mid C,U)}_{\text{residual variation}}.
    \label{eq:external-decomposition}
\end{equation}
The first term captures externally supplied information exposed by $X$; the
second captures residual unpredictability even given $C$ and $U$.
Consequently, substantive critiques not recoverable from $C$ can increase the
score, but so can stochastic generation. Because external information may also
come from retrieval or tools, we interpret \method as a graded proxy for
\emph{context-relative informational contribution}, not as a provenance or
cognitive-effort detector.

\subsection{Practical score}
\label{sec:hint-selection}

For a practical implementation of Eq.~\eqref{eq:ideal-sc}, we fix a
disclosure budget and a hint-construction procedure
$q_\phi(x\mid T)$. For instance, Random Words reveals a fixed fraction
$r$ of the words of $T$. Given i.i.d.\ samples
$x_1,\ldots,x_N\sim q_\phi(\cdot\mid T)$, define the per-token gain
associated with a hint $x$ as
\begin{equation}
    \Delta_\theta(x;T\mid C)
    :=
    \frac{1}{n}
    \sum_{t=1}^{n}
    \left[
        \log P_\theta(y_t\mid C+x+y_{<t})
        -
        \log P_\theta(y_t\mid C+y_{<t})
    \right].
    \label{eq:score-average}
\end{equation}
We estimate the Self-Conditioning score by averaging over the sampled hints:
\begin{equation}
    s^\phi_\theta(T\mid C)
    :=
    \frac{1}{N}
    \sum_{j=1}^{N}
    \Delta_\theta(x_j;T\mid C).
    \label{eq:practical-score}
\end{equation}
Here $P_\theta$ is the proxy language model and $+$ denotes concatenation.
Hints are prepended with the marker ``Hints:''. In the peer-review setting,
$C$ consists of the full paper together with a generic reviewing instruction
(Appendix~\ref{app:prompts}), while $T$ is the review being assessed.

\paragraph{Interpretation.}
Self-conditioning improves the likelihood of any coherent text, so our
hypothesis is comparative: hints should provide a larger gain when a review
contains information not predictable from $C$. Appendix~\ref{app:deferred}
illustrates this token by token.

\paragraph{Hints.}
Unless stated otherwise, Random Words reveals $r=0.3$ of the words of $T$ in
their original order, averaged over $N=10$ draws. Alternative hint families are
defined in Appendix~\ref{app:hint_selection} and compared in
Appendix~\ref{app:mega_ablation}.

%% ############################ END REWRITTEN BLOCK ###########################

%% ############################################################################
%% ############################################################################
%% ###                                                                      ###
%% ###   REWRITTEN BLOCK -- SECTION 4 -- EXPERIMENTS
%% ###   Peer reviews only; lexical-statistics table moved to Appendix; three explicit research questions (RQ1-RQ3) replacing the two undefined ones.
%% ###                                                                      ###
%% ############################################################################
%% ############################################################################

\section{Experiments} \label{sec:experiment_details}

\subsection{Data and models} \label{ssec:datasets_models}

We use the \emph{IntelLabs AI Peer Review Detection Benchmark}
\citep{yu-etal-2026-is-your-paper}, containing human reviews and counterparts
generated by Claude-Sonnet-3.5, GPT-4o, Gemini-1.5-Pro, Llama-3.1-70B, and
Qwen-2.5-72B. Because our score requires the production context $C$, we match
reviews to papers through the OpenReview API;\footnote{\url{https://docs.openreview.net/reference/api-v2}}
see Appendix~\ref{app:parsing}. Our main subset, \qt{IntelLabs-500}, contains
500 ICLR 2019 human reviews and their five generated counterparts. We restrict
human reviews to pre-2022 submissions; results for other years and NeurIPS are
in Appendix~\ref{app:more_data}. We additionally generate \textit{FD} reviews
with GPT-5.4 and Gemini-3.1-Pro.

We construct the remaining regimes of \S\ref{sec:regimes} from this subset.
\textit{MP} reviews are human reviews polished by Llama-3.1-8B-Instruct or
Qwen3-8B, while \textit{PI} reviews are generated from $k$ of eight structured
human-review bullet points. Prompts and generation parameters are in
Appendix~\ref{app:prompts}; lexical statistics are reported in
Table~\ref{tab:reviewer_metrics} (Appendix~\ref{app:deferred}).

Unless stated otherwise, $P_\theta$ is Llama-3.1-8B-Instruct. Alternative
scoring models are studied in \S\ref{sec:ablations}.

\subsection{Evaluation protocol}
\label{ssec:evaluation_protocol}

We interpret $s_\theta^\phi(\cdot)$ as an approximation of the average per-token description-length
reduction obtained by conditioning on hints from $\mathcal X_\phi$, and study three questions.

\paragraph{RQ1 -- Separability.}
Does the score separate fully-delegated reviews from human reviews and
their machine-polished counterparts? Under the information-gain
interpretation of Section~\ref{sec:external-information}, hints should
provide a larger explanatory gain for reviews whose content contains
information supplied beyond $C$.
\textit{MP} reviews should score higher than \textit{FD} ones, as the former’s content 
remains externally derived.
We report score distributions as kernel density estimates and quantify separability with pairwise AUC,
treating human-written texts as the positive class: higher AUC means more separable from human writing,
lower AUC means closer to it.

\paragraph{RQ2 -- Graded external information.}
Does the score vary with the amount of information supplied to the
generator? Supplying more human-derived bullet points increases the
external information available during generation and should move the
resulting reviews towards the human score distribution. Accordingly, AUC
against human reviews should decrease along the PI ladder.

\textbf{RQ3 -- Robustness and failure modes.} Is the score stable where it should be, and where does it
break? We examine reviews that are genuinely human but uninformative, and fully-delegated reviews made
artificially surprising through decoding parameters (\S\ref{sec:edge_cases}).

%% ############################ END REWRITTEN BLOCK ###########################

%% ############################################################################
%% ############################################################################
%% ###                                                                      ###
%% ###   REWRITTEN BLOCK -- SECTION 5 -- RESULTS
%% ###   Essays and reference letters removed; the two baseline subsections merged into one; detector descriptions moved to Appendix; hint-selection paragraph reduced to peer reviews; terminology switched to external information.
%% ###                                                                      ###
%% ############################################################################
%% ############################################################################

\section{Results} \label{sec:results}

\subsection{Polished reviews score like human ones (RQ1)}
\label{ssec:separability}

\begin{figure}[htb]
    \centering
    \includegraphics[width=0.9\linewidth]{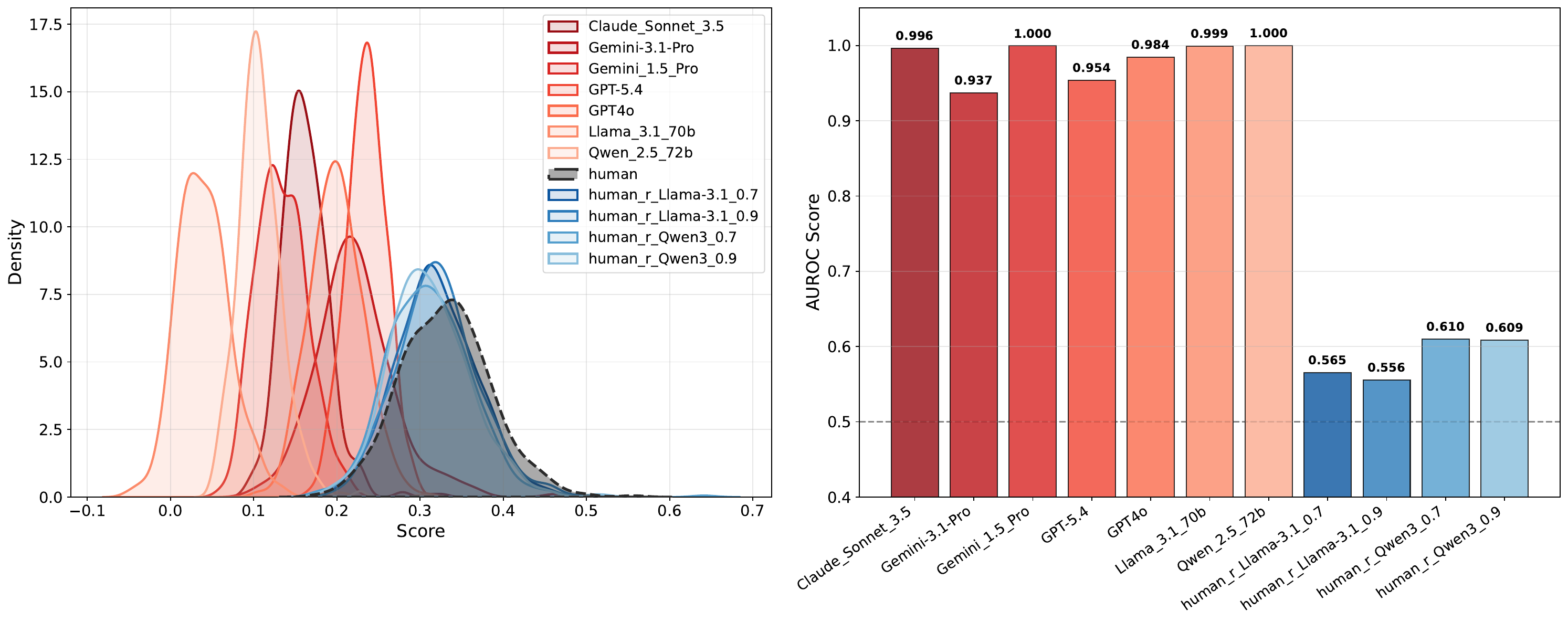}
    \caption{Distribution of \method scores for fully-delegated and machine-polished reviews (left), and
    their AUC with respect to human productions (right). \qt{human\_r\_\{model\_name\}\_T} labels
    rewritings of the human reviews by LLM \{model\_name\} at temperature T.}
    \label{fig:reviews_r_0.3}
\end{figure}

Figure~\ref{fig:reviews_r_0.3} shows that \method separates fully-delegated reviews from LLM-polished
human reviews. We consider both the artificial reviews released by
\citet{yu-etal-2026-is-your-paper} and our additional generations. Reviews from older or smaller models
remain easy to distinguish from human writing, whereas Gemini-3.1-Pro and GPT-5.4 overlap more with the
human distribution. Machine-polished reviews, in contrast, remain strongly aligned with human texts
(AUC between $0.556$ and $0.610$), even when the polishing model is small. This contrasts with
\citet{saha-feizi-2025-almost}, who show that such rewrites are routinely classified as fully
AI-generated by detectors, and supports our claim: polishing rewrites the encoding of a critique without
adding to it, so the external information of the review -- and hence its description-length profile --
is largely preserved.

\subsection{Standard detectors measure a different axis}
\label{ssec:other_methods}

\begin{table}[htbp]
    \centering
    \caption{AUC of baseline detectors across generator models, ICLR 2019 subset (N=500 per row).
        Left (\textit{vs. H}): human against fully-delegated reviews; right (\textit{vs. MP}):
        fully-delegated against machine-polished reviews. Detector details are in Appendix~\ref{app:baseline_details}.}
    \resizebox{\linewidth}{!}{
    \begin{tabular}{l|*{5}{rr|}rr}
    \toprule
    \textbf{Sample Generator} & \multicolumn{2}{c|}{\textbf{TMR-AI}} & \multicolumn{2}{c|}{\textbf{ModernBERT}} & \multicolumn{2}{c|}{\textbf{Binoculars}} & \multicolumn{2}{c|}{\textbf{Fast-DetectGPT}} & \multicolumn{2}{c|}{\textbf{EditLens}} & \multicolumn{2}{c}{\textbf{Ours}} \\
     \textit{Fully-delegated} & \textit{vs. H} & \textit{vs. MP} & \textit{vs. H} & \textit{vs. MP} & \textit{vs. H} & \textit{vs. MP} & \textit{vs. H} & \textit{vs. MP} & \textit{vs. H} & \textit{vs. MP} & \textit{vs. H} & \textit{vs. MP} \\
    \midrule
    \quad Claude-Sonnet-3.5 & 0.945 & 0.569 & 0.583 & 0.527 & 0.993 & 0.806 & 0.994 & 0.822 & 1.000 & 0.999 & 0.996 & 0.996 \\
    \quad GPT-4o            & 0.976 & 0.803 & 0.530 & 0.545 & 0.953 & 0.584 & 0.962 & 0.646 & 1.000 & 0.997 & 0.984 & 0.981 \\
    \quad Gemini-1.5-Pro    & 0.885 & 0.553 & 0.549 & 0.517 & 0.998 & 0.903 & 0.999 & 0.929 & 1.000 & 0.998 & 1.000 & 1.000 \\
    \quad Llama-3.1-70B     & 0.923 & 0.786 & 0.549 & 0.517 & 0.998 & 0.985 & 0.998 & 0.976 & 1.000 & 0.953 & 0.999 & 0.999 \\
    \quad Qwen2.5-72B       & 0.976 & 0.808 & 0.559 & 0.503 & 1.000 & 0.995 & 1.000 & 0.997 & 1.000 & 0.995 & 1.000 & 1.000 \\
    \quad GPT-5.4           & 0.642 & 0.798 & 0.641 & 0.801 & 0.546 & 0.961 & 0.546 & 0.952 & 1.000 & 0.940 & 0.954 & 0.941 \\
    \quad Gemini-3.1-Pro    & 0.508 & 0.585 & 0.507 & 0.587 & 0.825 & 0.644 & 0.774 & 0.796 & 0.978 & 0.709 & 0.937 & 0.922 \\
    \bottomrule
\end{tabular}}

    \label{tab:baseline_detectors_fd_mp}
\end{table}

Detection methods that key on grammatical and lexical patterns should be unable to distinguish
fully-delegated reviews (forbidden) from machine-polished ones (permitted). We test this with five
detectors: a ModernBERT variant \citep{warner-etal-2025-smarter} and a RoBERTa variant (TMR), both
fine-tuned on RAID \citep{dugan-etal-2024-raid} and MAGE \citep{li-etal-2024-mage} and the top-performing
open-source entries on the RAID leaderboard; the unsupervised Binoculars
\citep{hans-etal-2024-SpottingLLMsBinoculars} and Fast-DetectGPT \citep{bao-etal-2024-fastdetectgpt};
and EditLens \citep{thai-etal-2026-editlens}, which quantifies the extent of AI editing. Full
descriptions and settings are in Appendix~\ref{app:baseline_details}; results are in
Table~\ref{tab:baseline_detectors_fd_mp}.

For the older benchmark generators, most baselines distinguish fully-delegated
reviews from human ones, while performance deteriorates substantially for
GPT-5.4 and Gemini-3.1-Pro. The comparison with machine-polished reviews
reveals a more fundamental difference in what these methods measure.
Binoculars and Fast-DetectGPT confuse some fully-delegated generations with
machine-polished reviews, while TMR is similarly sensitive to surface
characteristics.

EditLens is an important exception: it separates fully-delegated from
machine-polished reviews well, as expected from a model explicitly trained
to quantify AI-assisted editing. However, this is not the quantity targeted
here. Machine-polished reviews may receive high AI-editing scores even when
their substantive critique is entirely human-derived. Conversely, as shown
in Section~\ref{ssec:graded_effort}, EditLens remains essentially saturated
across the partially-informed ladder, whereas Self-Conditioning changes
systematically with the amount of external information supplied. Thus our
claim is not that existing detectors can never separate these classes, but
that they do not represent the context-relative informational quantity
required by our formulation.

%% ============================================================
%% >>> ADDED EXPERIMENT -- polishing levels

\paragraph{Polishing intensity.}
Machine-polished reviews score like human ones, but one might object that this only shows the score is
insensitive, because polishing changes little. We therefore vary the intensity of the rewriting,
prompting Llama-3.1-8B-Instruct to edit each human review at increasing degrees, from light
copy-editing to full rephrasing (prompts in Appendix~\ref{app:prompts}).
Table~\ref{tab:rewriting_metrics} reports how far each output moved from its original and how separable
it remains from human reviews. The surface divergence is substantial: mean edit distance grows from
$289$ to $1843$ characters, i.e.\ from $11\%$ to $70\%$ of the average review length, while BLEU falls
from $0.74$ to $0.15$. Semantic distance, in contrast, stays below $0.14$ throughout, confirming that
the critique itself survives the rewriting. Our score follows the semantics rather than the surface:
AUC against human reviews rises only from $0.512$ to $0.561$, close to the $0.5$ chance level even at
the most aggressive setting. The residual increase indicates that very heavy rewriting does erode a
small amount of the measured external information, but an order of magnitude less than the surface
change alone would suggest.
 
\begin{table}[htbp]
\centering
\caption{Effect of rewriting intensity on machine-polished reviews ($N=500$ per row, rewriting model
Llama-3.1-8B-Instruct). Levenshtein, SBERT and BLEU are computed against the original human review;
\textit{AUC vs.\ H} is the separability of the polished reviews from human ones under \method, where
$0.5$ means indistinguishable.}
\label{tab:rewriting_metrics}
\begin{tabular}{lcccc}
\toprule
\textbf{Rewriting Level} & \textbf{Levenshtein Distance} & \textbf{SBERT Distance} & \textbf{BLEU Score} & \textbf{AUC} \textit{vs. H} \\
\midrule
Light Rewrite    & 289.3  & 0.0207 & 0.7361 & 0.512 \\
Moderate Rewrite & 1442.5 & 0.0840 & 0.2584 & 0.513 \\
Heavy Rewrite    & 1735.7 & 0.1164 & 0.1849 & 0.557 \\
Major Rewrite    & 1843.4 & 0.1329 & 0.1479 & 0.561 \\
\bottomrule
\end{tabular}
\end{table}

%% ============================================================

\subsection{The score is graded, not binary (RQ2)}
\label{ssec:graded_effort}

\begin{figure}[htb]
    \centering
    \includegraphics[width=0.9\linewidth]{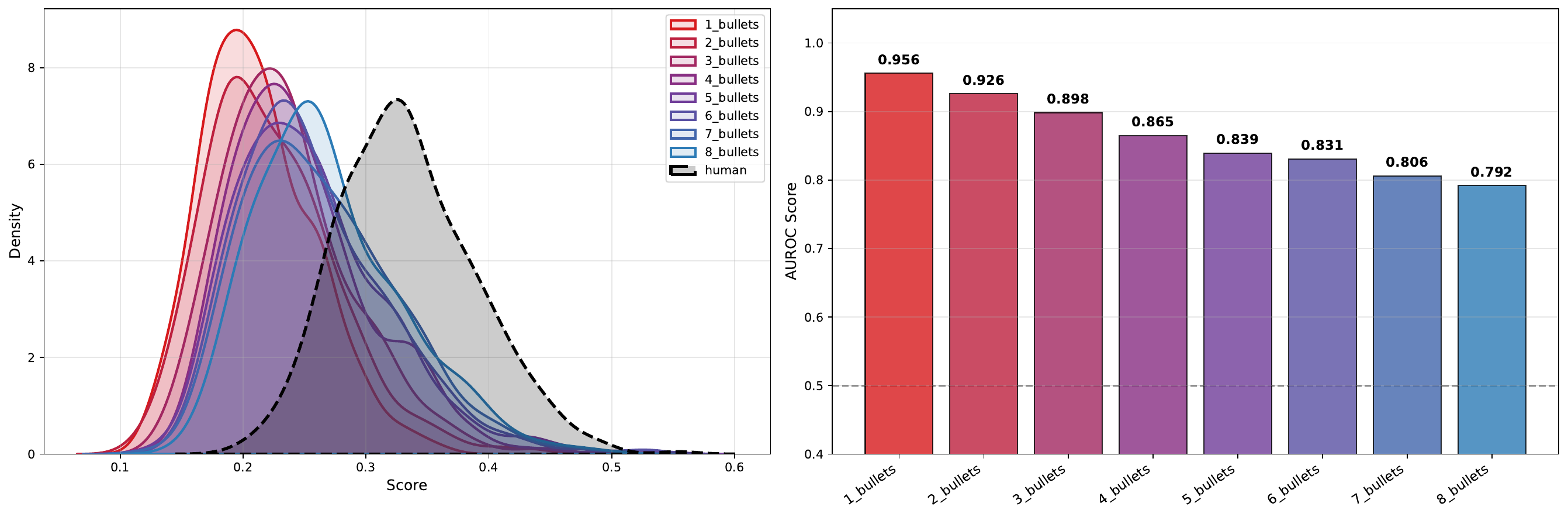}
    \caption{Reviews generated with Qwen3.5-35B-A3B from a varying number of bullet points supplied in
    the prompt. Left: distribution of scores; right: AUC with respect to human texts.}
    \label{fig:qwen_bullets}
\end{figure}

\begin{table}[htbp]
    \centering
    \caption{AUC of other detectors on reviews generated with Qwen3.5-35B-A3B from a varying number of
    bullet points supplied in the prompt.}
    \begin{tabular}{l|cccccccc}
        \textbf{Number of bullet points} & 1 & 2 & 3 & 4 & 5 & 6 & 7 & 8    \\
        \midrule
        \quad \textbf{Ours}                   & 0.96 & 0.93 & 0.90 & 0.87 & 0.84 & 0.83 & 0.81 & 0.80 \\
        \midrule
        \quad EditLens                        & 1.00 & 1.00 & 1.00 & 1.00 & 1.00 & 1.00 & 1.00 & 1.00 \\
        \quad Fast-DetectGPT                   & 0.79 & 0.80 & 0.84 & 0.83 & 0.84 & 0.84 & 0.85 & 0.85 \\
        \quad Binoculars                      & 0.78 & 0.79 & 0.83 & 0.83 & 0.84 & 0.84 & 0.84 & 0.85 \\
        \quad Modernbert(RAID/MAGE)           & 0.55 & 0.54 & 0.54 & 0.53 & 0.56 & 0.56 & 0.55 & 0.53 \\
        \bottomrule
    \end{tabular}
    \label{tab:other_detectors_perf}
\end{table}

To study graded levels of externally-provided information, we compress each review of
\qt{IntelLabs-500} into 8 structured bullet points with Llama-3.1-8B-Instruct, each capturing a distinct
aspect of the review (contribution, strengths, weaknesses, novelty, clarity, and so on), and regenerate
reviews conditioned on the paper plus a subset of these bullets. Increasing the number of bullets
increases the information budget available to the generator. As shown in Figure~\ref{fig:qwen_bullets}
with Qwen3.5-35B-A3B as generator, the human-versus-generated AUC decreases linearly ($r=-0.99$) as more
bullets are supplied: the generated reviews become progressively less distinguishable from human ones.
Results for other generators are in Appendix~\ref{app:more_bullet_points}.

Table~\ref{tab:other_detectors_perf} contrasts this behaviour with standard detectors on the same ladder.
They behave exactly as their design predicts: EditLens labels every hybrid review as 100\% AI, while
Binoculars and Fast-DetectGPT treat them as strongly machine-like under all prompt conditions (see Appendix~\ref{app:bullets-other-detectors} Figures~\ref{fig:editlens-bullets}, \ref{fig:bino-bullets} and \ref{fig:fast-bullets} respectively). From a
traditional ATD perspective, this is not a failure: these methods are simply not built to register a gradual
increase in external information. ModernBERT is a different case, remaining near chance throughout, which
indicates poor out-of-domain generalization. Self-Conditioning is the only
method here that exhibits the intended graded behaviour: its score
distributions move progressively towards the human regime as more external
information is supplied, and the corresponding AUC against human reviews
decreases monotonically.

%% ============================================================
%% >>> ADDED EXPERIMENT -- randomised bullet order (control, not a second headline)
%%   - Same ladder but sampling k of the 8 bullets uniformly at random per review.
%%   - Rules out the confound that the AUC decrease tracks WHICH aspects are disclosed rather than
%%     HOW MANY.
\paragraph{Randomised bullet order.}
In the ladder above, the prompted $k$ bullets always follow a fixed schema (summary,
strengths, weaknesses, and so on), so the observed decrease in AUC could reflect \emph{which} aspects
were disclosed rather than \emph{how many}. We therefore repeat the experiment with the same generator,
sampling $k$ of the $8$ bullets uniformly at random for each review.
Figure~\ref{fig:qwen_bullets_random} shows the same monotone decrease as the fixed-order values of Figure~\ref{fig:qwen_bullets}. 
What matters is how much the generator is told, suggesting that
the amount of external information is what drives the score.

%% ============================================================

\subsection{Alternative hint families}
We compare the alternative hint families of Section~\ref{sec:hint-selection}
in Appendix~\ref{app:hint_selection}. Keywords and Surprisal Spans produce
the largest dynamic range along the bullet-point ladder, while Random Words
and Random $n$-grams preserve the same overall trend. Prefix/Suffix
disclosure is less sensitive, and Top-ranked Sentences remains close to
chance. We use Random Words as the default because its selection procedure
is simple, local to the assessed review, and requires neither corpus-level
TF--IDF fitting nor scorer-dependent hint selection. Its stochastic
averaging does, however, incur additional inference cost; computational
trade-offs are discussed in Appendix~\ref{app:complexity}.

%% ############################ END REWRITTEN BLOCK ###########################

%% ############################################################################
%% ############################################################################
%% ###                                                                      ###
%% ###   REWRITTEN BLOCK -- SECTION 6 -- EDGE CASES
%% ###   Condensed; LazyReview figure moved to Appendix; terminology updated; placeholder kept for the quantitative degradation experiment.
%% ###                                                                      ###
%% ############################################################################
%% ############################################################################

\section{Edge Cases: Uninformative Human and Surprising LLM Reviews} \label{sec:edge_cases}

\method identifies reviews whose content is predictable from the paper and a generic reviewing
instruction, which raises two edge cases. First, a genuinely human but low-effort review may contain
little beyond superficial observations, and would score low not because it was delegated but because it
is uninformative. Second, a fully-delegated review generated to be surprising, through
prompting or decoding parameters, might score high with no external information at all.

\paragraph{Uninformative human reviews.}
We evaluate \method on the LazyReview dataset \citep{purkayastha-etal-2025-lazyreview}, which annotates
peer reviews for unconstructive reviewing behaviour, together with fully-delegated counterparts we
generated by retrieving the original papers and following the procedure of
\S\ref{sec:experiment_details} with Llama-3.1-70B-IT. As shown in Figure~\ref{fig:lazy_reviews}
(Appendix~\ref{app:deferred}), the human reviews remain well separated from delegated ones ($1.0$ AUC). This result
should be read with care: a review is labelled \qt{lazy} as soon as one of its sentences exhibits
lazy-thinking, while the rest of the review may be perfectly informative.

\paragraph{Surprising machine texts.}
Figure~\ref{fig:high_temp_reviews} compares human ICLR 2019 reviews with
fully-delegated reviews generated by Llama-3.1-70B-Instruct at temperatures
$1.0$, $1.5$, and $2.0$. Increasing temperature moves machine scores toward
the human range: discrimination remains strong at $1.0$ (perfect AUC) and
$1.5$ (AUC $0.926$), while at $2.0$ the generated reviews score above their
human counterparts. This failure mode follows from
Eq.~\eqref{eq:external-decomposition}: although
$I(U;X\mid C)=0$ for fully-delegated reviews, increased sampling randomness
can inflate the residual term $I(T;X\mid C,U)$.

This evasion comes at a cost. As shown in
Appendix~\ref{app:high-temp}, reviews sampled at $T=2.0$ exhibit substantial
degradation in coherence, grammatical, and logical consistency, making them conspicuous to
human readers. Moreover, such temperatures are atypical for analytical
generation and are unavailable in some current systems with fixed sampling
settings. Since Self-Conditioning is intended as a screening tool rather than
a standalone decision rule, such low-quality false negatives can still be
identified during human inspection.

%% ############################ END REWRITTEN BLOCK ###########################

%% ############################################################################
%% ############################################################################
%% ###                                                                      ###
%% ###   REWRITTEN BLOCK -- SECTION 7 -- ABLATIONS
%% ###   Both ablation tables moved to the Appendix; findings summarised in one paragraph each.
%% ###                                                                      ###
%% ############################################################################
%% ############################################################################

\section{Ablations} \label{sec:ablations}

\paragraph{Disclosure budget.}
The score depends on the fraction $r$ of the review revealed as hints.
Varying $r\in\{0.2,0.3,0.4\}$ (Table~\ref{tab:other_ratio_perf}) shows no
uniformly optimal value. For comparisons against human reviews, all older
benchmark generators remain at or above $0.98$ AUC across the tested
ratios, whereas performance on GPT-5.4 and Gemini-3.1-Pro is more sensitive
to $r$. The disclosure budget is therefore stable for the older generators
but remains a tunable component for newer models.

\paragraph{Scoring model.}
All main experiments use Llama-3.1-8B-Instruct as $P_\theta$. Table~\ref{tab:other_scoring_perf}
(Appendix~\ref{app:deferred}) compares it with Llama-3.2-1B-Instruct, Llama-3.1-70B-Instruct and
Qwen3-8B. Our results show that no model is uniformly best:
the 70B Llama is worse than the 8B one on Gemini-generated reviews, and Qwen3-8B is better on Gemini but
worse on GPT-5.4. The 1B variant is the weakest overall. Scoring with Llama-3.1-70B-Instruct
also covers the special case $P_\theta = P_G$, where our method reaches perfect separation, but 
such scores are also reached for other generators, confirming that this setting is favourable
but not necessary.

%% ############################ END REWRITTEN BLOCK ###########################

%% ############################################################################
%% ############################################################################
%% ###                                                                      ###
%% ###   REWRITTEN BLOCK -- SECTION 8 -- CONCLUSION AND LIMITATIONS
%% ###   Reframed on external information; the source-agnosticism limitation added; temperature moved from unknown to characterised; multi-domain caveat rewritten for a peer-review-only paper.
%% ###                                                                      ###
%% ############################################################################
%% ############################################################################

\section{Summary and Concluding Remarks}
\label{sec:conclusion}

We introduced a new perspective on LLM-assisted peer reviewing: rather than measuring how much of a text
was written by a machine, we estimate how much \emph{external information} it carries beyond its
production context. The resulting score is unsupervised, requires no annotated data, and is
computed from a single open-weights proxy model. Empirically it behaves as a measure of external
information. 
Machine-polished human reviews remain close to the human
Self-Conditioning distribution even after substantial rewriting. Standard
ATD methods instead primarily reflect machine involvement in the observed
text and may therefore assign high AI-editing scores to such reviews.
Along a controlled ladder of externally-provided bullet points,
Self-Conditioning varies systematically with the amount of information
supplied to the generator, converging towards the human regime. Together,
these results indicate that context-relative informational contribution is
a distinct axis from conventional AI-text detection.

\paragraph{Limitations.}
\method should not be used as a standalone misconduct detector. A low score indicates that a review
contains little information beyond what the scoring model can infer from the paper and prompt; it does
not prove that the review was generated by an LLM. \method is therefore best viewed as a
decision-support tool for identifying reviews that warrant further human inspection.

By design, Self-Conditioning measures information beyond the declared
context, not its provenance. Although the external information in our
controlled experiments is human-derived, retrieval, external tools, or
other sources could produce the same effect. A high score therefore does
not establish human authorship, just as a low score does not establish
delegation.

Absolute scores depend on the scoring model, context construction, hint mechanism, and evaluated text distribution; comparisons should therefore be made within a fixed experimental setting.
High-temperature sampling is a
characterised failure mode rather than an open question (\S\ref{sec:edge_cases}), but two related
scenarios remain untested: fully-delegated reviews later \qt{humanised} by users, for which
we could find no paired review-paper data, and adversarial prompting designed to spike perplexity
without the user reading the paper. Finer-grained mixing, where human and machine content alternate
sentence by sentence, is also not covered. Finally, all our data consist of ML papers and their reviews written in English, 
which does not permit a general assessment of robustness and generalisation.

%% ############################ END REWRITTEN BLOCK ###########################

%% ============================================================

\todos

\bibliographystyle{plainnat}
\bibliography{custom, anthology-1, anthology-2}

\appendix

%% ############################################################################
%% ###   NEW APPENDIX SECTION -- material moved out of the main text          ###
%% ############################################################################

\section{Additional Theoretical Properties of Self-Conditioning}
\label{app:theory}
\subsection{Coding interpretation}
\label{app:coding-interpretation}

Equation~\eqref{eq:conditional-information-density} has a direct coding
interpretation. Under an ideal probabilistic code, the description lengths
of $t$ with and without $x$ are respectively
\[
    \ell(t\mid c,x)=-\log P(t\mid c,x),
    \qquad
    \ell(t\mid c)=-\log P(t\mid c).
\]
Hence
\begin{equation}
    \imath(t;x\mid c)
    =
    \ell(t\mid c)-\ell(t\mid c,x),
    \label{eq:coding-gain}
\end{equation}
so that the information density is exactly the number of nats saved when
$x$ is made available. Thus the description-length intuition underlying
our criterion follows directly from the one-shot likelihood ratio, without
requiring an optimization over a latent space of possible explanations.

\subsection{Likelihood-ratio interpretation}
\label{app:hypothesis-testing}

The likelihood ratio in
\eqref{eq:conditional-information-density} also admits a one-shot
hypothesis-testing interpretation. Condition on $C=c$ and consider the two
hypotheses
\begin{align}
    H_1 &: (T,X)\sim P_{TX\mid C=c},
    \label{eq:hyp-dependent}\\
    H_0 &: (T,X)\sim
        P_{T\mid C=c}P_{X\mid C=c}.
    \label{eq:hyp-independent}
\end{align}
Under $H_1$, the disclosed fragment and the text have their genuine joint
distribution; under $H_0$, the disclosure contains no text-specific
information beyond that already provided by $c$. Their log-likelihood ratio
is
\begin{align}
    \log
    \frac{P_{TX\mid C}(t,x\mid c)}
         {P_{T\mid C}(t\mid c)P_{X\mid C}(x\mid c)}
    &=
    \log
    \frac{P(t\mid c,x)}
         {P(t\mid c)}
    \nonumber\\
    &=
    \imath(t;x\mid c).
    \label{eq:llr-information-density}
\end{align}
Thus the quantity underlying Self-Conditioning is precisely the
likelihood-ratio statistic for distinguishing a genuine text--disclosure
pair from one in which the disclosure carries no additional information
about the text beyond the baseline context. This provides a one-shot
operational meaning to the score independently of any asymptotic
interpretation.

\subsection{Population interpretation}
\label{app:population-interpretation}

Although our primary object is the information gain of an individual text,
its expectation has a familiar information-theoretic interpretation. Let
$(C,T)\sim P_{CT}$ and $X\sim q_\phi(\cdot\mid T)$. Then
\begin{align}
    \mathbb{E}\!\left[
        \log
        \frac{P(T\mid C,X)}
             {P(T\mid C)}
    \right]
    &=
    \mathbb{E}\!\left[
        \log
        \frac{P_{TX\mid C}(T,X\mid C)}
             {P_{T\mid C}(T\mid C)P_{X\mid C}(X\mid C)}
    \right]
    \nonumber\\
    &=
    I(T;X\mid C).
    \label{eq:sc-cmi}
\end{align}
The first equality follows from
\[
    P(T\mid C,X)
    =
    \frac{P_{TX\mid C}(T,X\mid C)}
         {P_{X\mid C}(X\mid C)}.
\]

Consequently,
\begin{equation}
    \mathbb{E}[\mathcal S_\phi(T\mid C)]
    =
    \mathbb{E}\!\left[
        \frac{1}{|T|}
        \imath(T;X\mid C)
    \right].
    \label{eq:sc-pop-length}
\end{equation}
When all texts have common length $n$, this simplifies to
\begin{equation}
    \mathbb{E}[\mathcal S_\phi(T\mid C)]
    =
    \frac{1}{n}I(T;X\mid C).
    \label{eq:sc-pop}
\end{equation}

Self-Conditioning can therefore be viewed at two complementary levels.
At the instance level, it averages one-shot information densities over
randomised disclosures of the particular text being assessed. At the
population level, the corresponding unnormalised expected gain is exactly
a conditional mutual information.

\subsection{Decomposition into external contribution and residual variation}
\label{app:external-decomposition}

Recall the Markov relation
\[
    (C,U)\longrightarrow T\longrightarrow X,
\]
which follows from the fact that the disclosure $X$ is constructed
exclusively from $T$. By the chain rule for conditional mutual information,
\begin{align}
    I(T,U;X\mid C)
    &=
    I(U;X\mid C)
    +
    I(T;X\mid C,U),
    \label{eq:chain-rule-u-first}\\
    I(T,U;X\mid C)
    &=
    I(T;X\mid C)
    +
    I(U;X\mid C,T).
    \label{eq:chain-rule-t-first}
\end{align}
The Markov relation implies
\[
    I(U;X\mid C,T)=0.
\]
Combining the two chain-rule expansions therefore gives
\begin{equation}
    I(T;X\mid C)
    =
    I(U;X\mid C)
    +
    I(T;X\mid C,U),
\end{equation}
which is Equation~\eqref{eq:external-decomposition} in the main text.

The first term measures the part of the latent information supplied beyond
$C$ that is recoverable through the disclosure $X$. The second term
captures information arising from aspects of the realised text that remain
unpredictable even when both $C$ and $U$ are known. This residual can arise,
for example, from stochastic generation or idiosyncratic realization
choices. Consequently, Self-Conditioning measures context-relative
informational contribution rather than provenance: external contribution
can increase the score, but unpredictability unrelated to external
information can do so as well.

\subsection{Monotonicity under scoring-context augmentation}
\label{app:context-augmentation}

The formulation gives an exact monotonicity result when information that
was previously external is made available in the conditioning context for
a fixed target distribution. Suppose that some external information $U$ is
added to the baseline context. The residual Self-Conditioning information
then becomes $I(T;X\mid C,U)$. From
Equation~\eqref{eq:external-decomposition},
\begin{equation}
    I(T;X\mid C)-I(T;X\mid C,U)
    =
    I(U;X\mid C)
    \geq 0.
    \label{eq:context-monotonicity}
\end{equation}
Thus information that is moved from outside the conditioning context into
the declared context can only reduce the residual Self-Conditioning
information, and the reduction is exactly the amount of that information
recoverable through the disclosure.

More generally, for a sequence of contributions
$U_1,\ldots,U_K$, define
\[
    C_k=(C,U_1,\ldots,U_k).
\]
Applying the same decomposition at each step gives
\begin{equation}
    I(T;X\mid C_{k-1})
    -
    I(T;X\mid C_k)
    =
    I(U_k;X\mid C_{k-1})
    \geq 0,
    \label{eq:nested-monotonicity}
\end{equation}
and hence
\begin{equation}
    I(T;X\mid C)
    \geq
    I(T;X\mid C_1)
    \geq \cdots \geq
    I(T;X\mid C_K).
    \label{eq:nested-ordering}
\end{equation}
This formalizes the context-relative nature of the criterion: information
counts as external only to the extent that it is not already included in
the conditioning context.

Importantly, this result concerns \emph{scoring-context augmentation}: the
target distribution is held fixed while progressively more information is
included in the context used to explain $T$. It is distinct from the
partially-informed experiments in Section~\ref{ssec:graded_effort}, where the
amount of information supplied during the \emph{generation} of $T$ is
varied while the baseline scoring context is held fixed. The monotonic
trend observed in those experiments is therefore an empirical property of
the proposed score, rather than a direct consequence of
Equation~\eqref{eq:nested-ordering}.

\FloatBarrier
\section{Additional Figures and Tables} \label{app:deferred}

\begin{figure}[htbp]
    \centering
    \includegraphics[width=\linewidth]{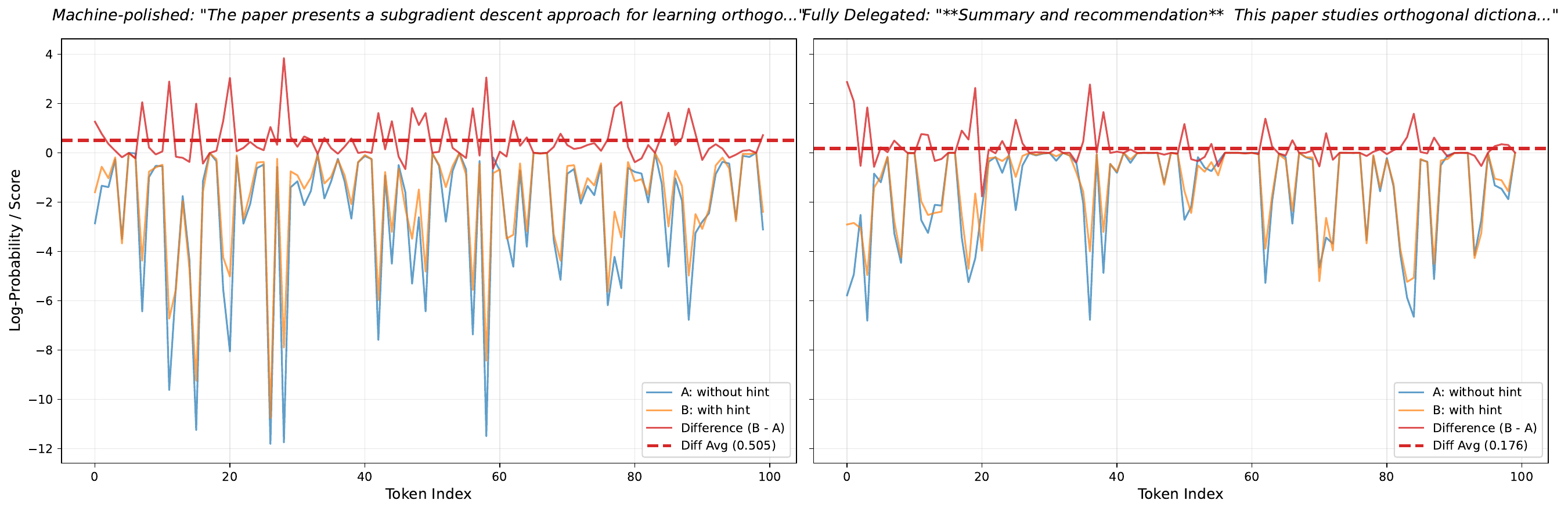}
    \caption{Variations of the token-level log-probability, without and with additional
    self-conditioning hints, and of the token-level log-likelihood ratio increment. The
    machine-polished review (left) has a larger average value than the fully-delegated one (right).}
    \label{fig:token_level_difference}
\end{figure}

\paragraph{Token-level illustration.}
Figure~\ref{fig:token_level_difference} contrasts a machine-polished and a fully-delegated review of the
same paper. It plots, token by token, the log-probability of the review under the context alone and
under the context augmented with self-conditioning hints, together with their difference. The polished
review receives a markedly larger average boost ($0.505$ against $0.176$), which is the effect
Eq.~\eqref{eq:score-average} aggregates.

\paragraph{Lexical statistics of the review sources.}
Table~\ref{tab:reviewer_metrics} details surface-level characteristics across all sources. A primary
distinction between human and fully-delegated reviews lies in their structure: human critiques vary
drastically in length (standard deviation of $284.7$ words) whereas generations from the IntelLabs
models are standardised and uniform. When models are asked to polish existing human reviews, they
inherit the human length distribution. GPT-5.4 is extremely verbose compared with the human baseline,
even when expanding short critiques, while Gemini-3.1-Pro behaves oppositely and prefers shorter
reviews; the two are very different from the IntelLabs generations and from one another, despite both
being reasoning models. Length alone therefore does not explain the score distributions reported in
\S\ref{sec:results}.

\begin{table}[htbp]
\centering
\caption{Lexical and statistical metrics for each review type. Each row corresponds to 500 reviews.}
\resizebox{\columnwidth}{!}{
\begin{tabular}{lrrrrr}
\toprule
\multirow{2}{*}{\textbf{Reviewer}} & \multicolumn{2}{c}{\textbf{Char count}} & \multicolumn{2}{c}{\textbf{Word count}} & \multirow{2}{*}{\textbf{Avg TTR}} \\
\cmidrule(lr){2-3} \cmidrule(lr){4-5}
& \textbf{Avg} & \textbf{Std} & \textbf{Avg} & \textbf{Std} & \\
\midrule
\textbf{Human}              & 2615.7 & 1709.7 & 428.0 & 284.7 & 0.528 \\
\midrule
\multicolumn{6}{l}{\textit{\textbf{IntelLabs Dataset}}}      \\
GPT-4o              & 2992.2 & 370.2 & 412.6 & 52.0 & 0.551 \\
Gemini-1.5-Pro & 2548.9 & 349.9 & 363.0 & 50.9 & 0.533 \\
Claude-Sonnet-3.5 & 2205.6 & 204.0 & 298.9 & 29.7 & 0.588 \\
Llama-3.1-70B & 1837.7 & 553.4 & 277.4 & 84.5 & 0.496 \\
Qwen2.5-72B & 3300.0 & 580.9 & 481.0 & 87.2 & 0.452 \\
\midrule
\multicolumn{6}{l}{\textit{\textbf{Added Generations}}}      \\
GPT-5.4            & 7602.0 & 1350.3 & 1077.7 & 197.5 & 0.442 \\
Gemini-3.1-Pro & 1740.1 & 1685.7 & 249.3 & 241.9 & 0.678 \\
\midrule
\multicolumn{6}{l}{\textit{\textbf{Polished human reviews}}}      \\
Qwen3-8B (t=0.7) & 2785.4 & 1661.0 & 436.5 & 266.7 & 0.546 \\
Qwen3-8B (t=0.9) & 2790.8 & 1654.1 & 437.2 & 265.5 & 0.545 \\
Llama-3.1-8B-IT (t=0.7) & 2433.3 & 1150.7 & 377.9 & 186.0 & 0.549 \\
Llama-3.1-8B-IT (t=0.9) & 2446.2 & 1145.1 & 378.2 & 183.2 & 0.553 \\
\bottomrule
\end{tabular}}
\label{tab:reviewer_metrics}
\end{table}

\begin{figure}[htbp]
    \centering
    \includegraphics[width=0.85\linewidth]{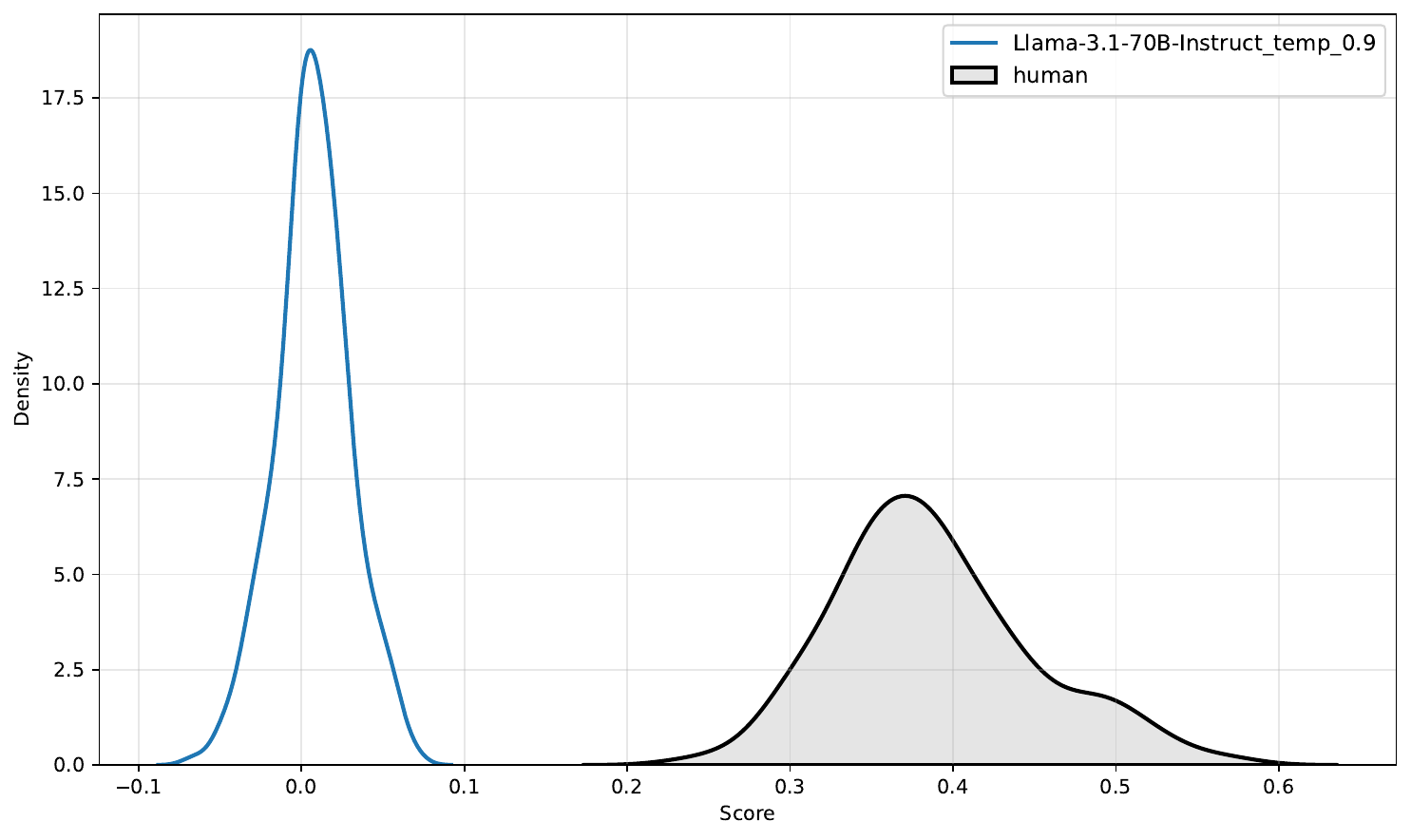}
    \caption{Distribution of scores for the \qt{Lazy Reviews} and their generated counterparts.}
    \label{fig:lazy_reviews}
\end{figure}

\paragraph{Uninformative human reviews.}
Figure~\ref{fig:lazy_reviews} accompanies \S\ref{sec:edge_cases}: it displays scores from the
LazyReview dataset \citep{purkayastha-etal-2025-lazyreview} together with the fully-delegated
counterparts we generated for the same papers.

\paragraph{Disclosure budget.}
Table~\ref{tab:other_ratio_perf} reports AUC for three values of the disclosure ratio $r$, discussed in
\S\ref{sec:ablations}.

\begin{table}[htbp]
    \centering
    \caption{AUC of our method across generator models in the ICLR 2019 subset for different disclosure
    ratios. $r=0.3$ is the ratio used in all other experiments.}
    \resizebox{\linewidth}{!}{
    \begin{tabular}{l|rr|rr|rr}
        \toprule
        \multirow{2}{*}{\textit{Fully-delegated}} & \multicolumn{2}{c|}{$\mathbf{r=0.2}$} & \multicolumn{2}{c|}{$\mathbf{r=0.3}$} & \multicolumn{2}{c|}{$\mathbf{r=0.4}$} \\
         & \textit{vs. H} & \textit{vs. MP} & \textit{vs. H} & \textit{vs. MP} & \textit{vs. H} & \textit{vs. MP} \\
        \midrule
        \quad Claude-Sonnet-3.5   & 0.993 & 0.989 & 0.996 & 0.996 & 0.998 & 0.997 \\
        \quad GPT-4o              & 0.988 & 0.981 & 0.984 & 0.981 & 0.980 & 0.969 \\
        \quad Gemini-1.5-Pro      & 0.999 & 0.998 & 1.000 & 1.000 & 1.000 & 1.000 \\
        \quad Llama-3.1-70B       & 0.999 & 0.998 & 0.999 & 0.999 & 1.000 & 1.000 \\
        \quad Qwen2.5-72B         & 1.000 & 0.999 & 1.000 & 1.000 & 1.000 & 1.000 \\
        \quad GPT-5.4             & 0.966 & 0.949 & 0.954 & 0.941 & 0.936 & 0.895 \\
        \quad Gemini-3.1-Pro      & 0.862 & 0.823 & 0.937 & 0.922 & 0.973 & 0.960 \\
        \bottomrule
    \end{tabular}%
    }
    \label{tab:other_ratio_perf}
\end{table}

\paragraph{Alternative scoring models.}
Table~\ref{tab:other_scoring_perf} reports AUC when $P_\theta$ is replaced by other open-weights models,
discussed in \S\ref{sec:ablations}.

\begin{table}[htbp]
    \centering
    \caption{AUC of our method across generators with various scoring models $P_\theta$ (in columns).
    The machine-polished texts are those in the bottom four rows of Table~\ref{tab:reviewer_metrics}.}
    \resizebox{\linewidth}{!}{%
    \begin{tabular}{l|rr|rr|rr}
        \toprule
        \multirow{2}{*}{\textit{Fully-delegated}} & \multicolumn{2}{c|}{\textbf{Llama-1B}} & \multicolumn{2}{c|}{\textbf{Llama-70B}} & \multicolumn{2}{c}{\textbf{Qwen3-8B}} \\
         & \textit{vs. H} & \textit{vs. MP} & \textit{vs. H} & \textit{vs. MP} & \textit{vs. H} & \textit{vs. MP} \\
        \midrule
        \quad Claude-Sonnet-3.5   & 0.890 & 0.810 & 0.997 & 0.997 & 0.997 & 0.991 \\
        \quad GPT-4o              & 0.934 & 0.879 & 0.986 & 0.983 & 0.986 & 0.955 \\
        \quad Gemini-1.5-Pro      & 0.976 & 0.951 & 1.000 & 1.000 & 0.998 & 0.994 \\
        \quad Llama-3.1-70B       & 0.995 & 0.993 & 1.000 & 1.000 & 0.999 & 0.995 \\
        \quad Qwen2.5-72B         & 0.994 & 0.990 & 1.000 & 1.000 & 1.000 & 0.999 \\
        \quad GPT-5.4             & 0.863 & 0.765 & 0.963 & 0.956 & 0.927 & 0.810 \\
        \quad Gemini-3.1-Pro      & 0.748 & 0.646 & 0.915 & 0.902 & 0.964 & 0.915 \\
        \bottomrule
    \end{tabular}%
    }
    \label{tab:other_scoring_perf}
\end{table}

\begin{figure}[htbp]
    \centering
    \includegraphics[width=0.7\linewidth]{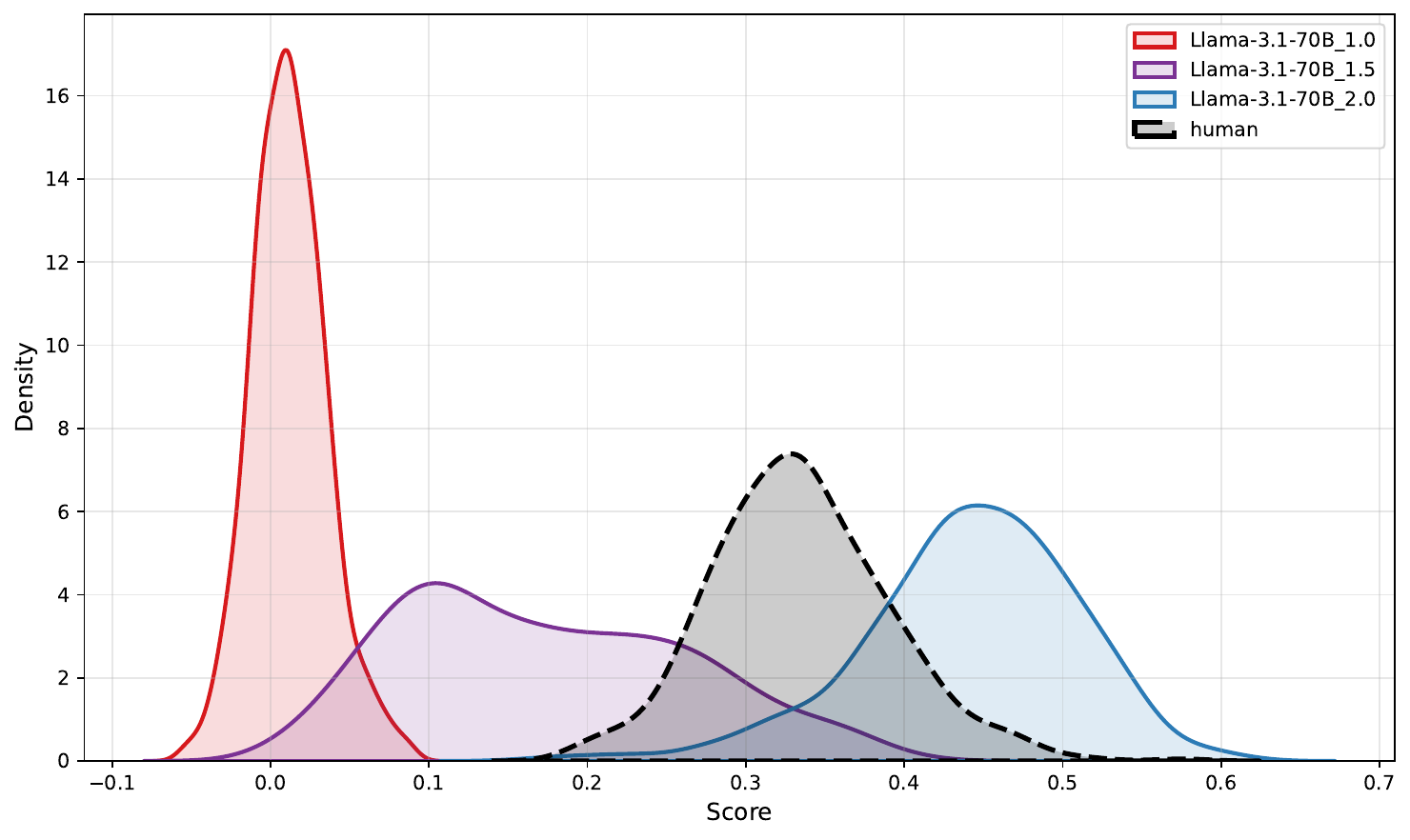}
    \caption{Scores for ICLR 2019 human reviews and high-temperature fully-delegated counterparts.}
    \label{fig:high_temp_reviews}
\end{figure}

\paragraph{Surprising LLM Reviews}
Fig~\ref{fig:high_temp_reviews} displays the scores and respective AUC versus the human reviews of the generated counterparts with high temperature settings.

\paragraph{Other detectors densities on reviews generated from bullet points.}
\label{app:bullets-other-detectors}

\begin{figure}[htbp]
    \centering
    \includegraphics[width=0.9\linewidth]{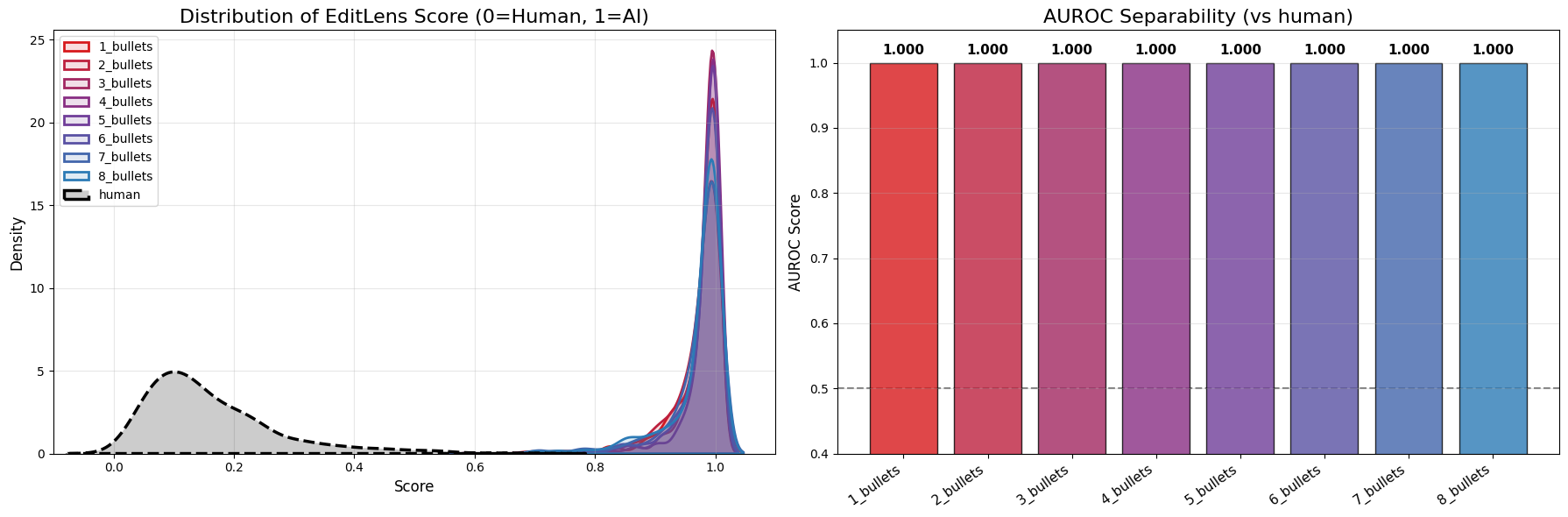}
    \caption{EditLens density of scores on the generated reviews from bullet points}
    \label{fig:editlens-bullets}
\end{figure}

\begin{figure}[htbp]
    \centering
    \includegraphics[width=0.9\linewidth]{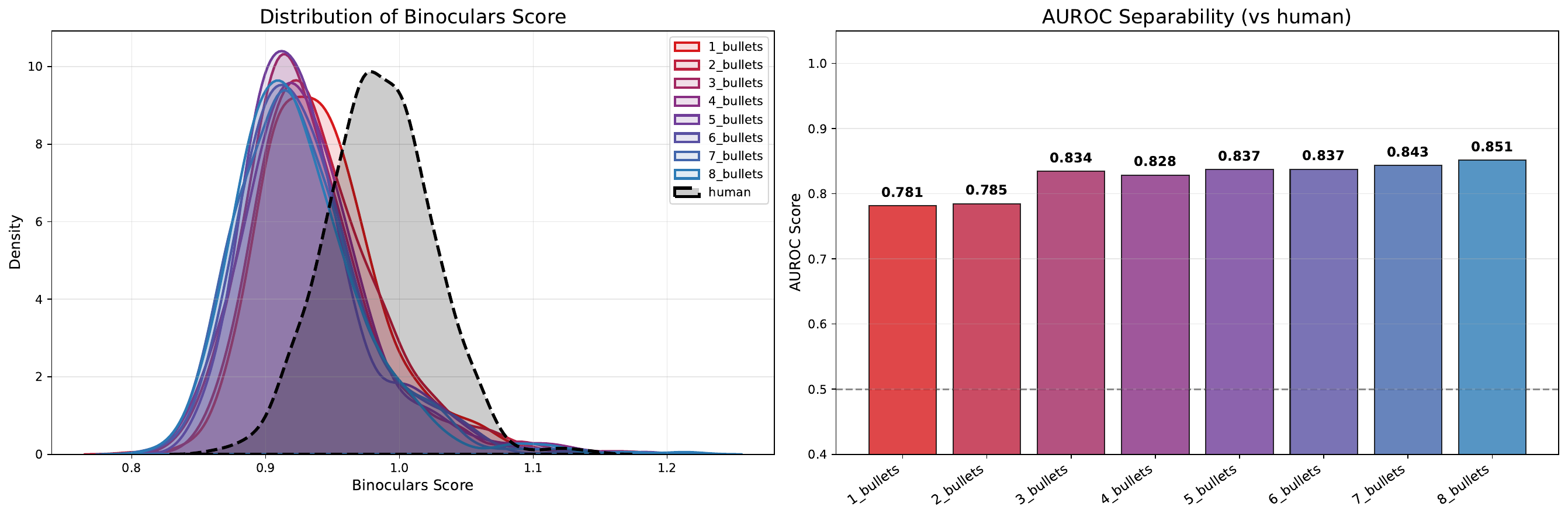}
    \caption{Binoculars density of scores on the generated reviews from bullet points}
    \label{fig:bino-bullets}
\end{figure}

\begin{figure}[htbp]
    \centering
    \includegraphics[width=0.9\linewidth]{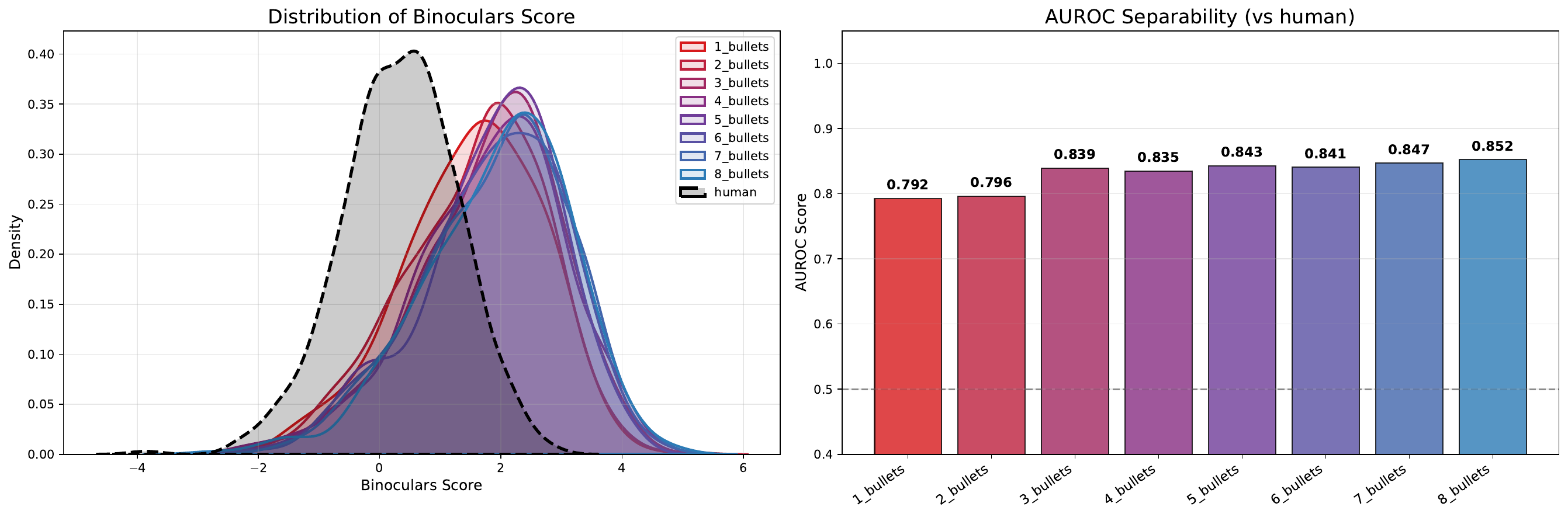}
    \caption{Fast-DetectGPT density of scores on the generated reviews from bullet points}
    \label{fig:fast-bullets}
\end{figure}

\begin{figure}[htbp]
    \centering
    \includegraphics[width=0.9\linewidth]{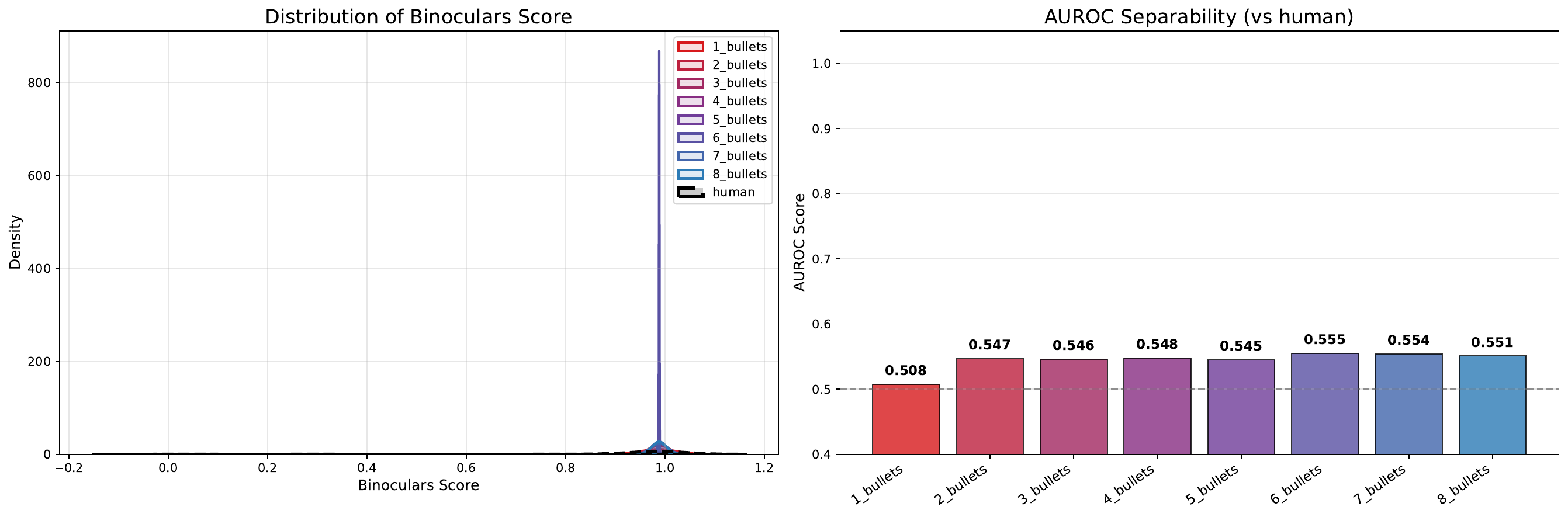}
    \caption{ModernBERT-based detector’s density of scores on the generated reviews from bullet points}
    \label{fig:modernbert-bullets}
\end{figure}

%% >>> SKELETON NOTE: RECONCILE: the base paper reports ~4000 GPU-hours (Appendix~\ref{app:complexity}) while the peer-review paper reported ~2000. Pick one, consistent with the experiments that ship.
\FloatBarrier
\section{Detailed List of Models Used}
\label{app:more_infos_models}

Unless specified otherwise, Llama-3.1-8B-Instruct is used as the scoring model $P_\theta$ in all main experiments. The prompts relative to all these generations and rewritings are available in Appendix Section~\ref{app:prompts}

\paragraph{Peer reviews.}
The \textbf{IntelLabs Dataset} used human reviews from OpenReview, we selected samples released before 2022 to avoid any AI influence. They also used 5 different models to generate their peer-reviews: Claude-Sonnet-3.5, GPT-4o, Gemini-1.5-Pro, Llama-3.1-70B and Qwen-2.5-72b.

To augment this dataset, we used GPT-5.4 and Gemini-3.1-Pro (in March 2026) to generate additional fully generated reviews, Llama-3.1-Instruct-8B and Qwen3-8B to produce rewritten human reviews, and Llama-3.1-Instruct-8B to summarize human reviews into structured bullet points. Hybrid reviews conditioned on these bullet-point summaries are then generated with Qwen3.5-35B-A3B, Gemma-4-E4B-it, and Mistral-Small-3.2-24B-Instruct-2506.
Rewritings were done using either a temperature of 0.7 or 0.9 with 2048 max new tokens. Summarisation was performed using greedy decoding. Fully-generated reviews were obtained with a temperature of 0.9 and 2048 completion tokens. Reviews generated from bullet points were also obtained using a temperature of 0.9.

% \paragraph{Essays.}
% The original Essay dataset we used includes ASAP-AES essays along with GPT-3.5 generations and rewritings. The exact version of GPT-3.5 used is unknown as it is not mentioned in the original work.

% For our augmentations, both fully generated texts and rewritten variants are produced with Mistral-Small-3.2-24B-Instruct-2506, Qwen3.5-35B-A3B, and Gemma-4-26B-A4B-it. All used a temperature of 0.9.

% \paragraph{Reference Letters.}
% The Reference Letters we used were all generated using ChatGPT. The exact version and timestamp of generation are not provided in the original work.

%% >>> SKELETON NOTE: IMPORTED FROM B: PDF parsing via the OpenReview API, which explains why the full IntelLabs dataset could not be used.
\FloatBarrier
\section{Parsing PDFs} \label{app:parsing}
Since Equation~\eqref{eq:score-average} requires to input the full paper content, we had to use the OpenReview API to match the reviews from the IntelLabs-AI-Peer-Review dataset and their corresponding papers. The API would return PDF files rather than pure text, so we used \verb|PyMuPDF4LLM| to convert those into markdown. Some images and figures caused the parsing to crash for some papers, and thus they were ignored along with their corresponding reviews, which explains why we could not use the whole dataset provided by \cite{yu-etal-2026-is-your-paper}.
\FloatBarrier
\section{Baseline Details} \label{app:baseline_details}

%% >>> MOVED HERE from the main text: full descriptions of the five baseline detectors.

\paragraph{Detectors evaluated.}
\textbf{A ModernBERT variant} \citep{warner-etal-2025-smarter}, fine-tuned on the RAID
\citep{dugan-etal-2024-raid} and MAGE \citep{li-etal-2024-mage} corpora, available at
\url{https://huggingface.co/GeorgeDrayson/modernbert-ai-detection-raid-mage}.
\textbf{A RoBERTa variant}, dubbed TMR (Text-Mining RoBERTa), also fine-tuned on
RAID.\footnote{\url{https://huggingface.co/Oxidane/tmr-ai-text-detector}} These two are the
top-performing open-source detectors on the RAID \ATD{} leaderboard, published at COLING~2025 and
regularly updated since.
\textbf{Binoculars} \citep{hans-etal-2024-SpottingLLMsBinoculars}, whose detection score is the quotient
of the average surprisal of the current token under a \emph{main model} $q$ over the average
cross-entropy between an \emph{auxiliary model} $r$ and $q$ at each token step; we use the default
setting, falcon-7b and its instruct version \citep{almazrouei-etal-2023-FalconSeriesOpen}. It is the
best-performing unsupervised method reported by \citet{yu-etal-2026-is-your-paper}.
\textbf{Fast-DetectGPT} \citep{bao-etal-2024-fastdetectgpt} follows the same intuition, computing the
normalised difference between the surprisal of the current token under $q$ and the empirical
cross-entropy between $q$ and another model $r$, averaged over the text.
\textbf{EditLens} \citep{thai-etal-2026-editlens} is a supervised approach quantifying the extent of
AI-assisted editing rather than assigning a binary label: a text scored $0.70$ is deemed to have been
70\% edited by an LLM.

Both Binoculars and Fast-DetectGPT were run with Falcon-7b and its instruct version as underlying models. Those two are the default configuration for Binoculars, and it is also the best combination for Fast-DetectGPT as argued by the authors themselves in their subsequent work \citep{bao-etal-2025-decoupling}. The results of these two methods are very similar because despite the intuitions being different, the final mathematical operations are the same as pointed out by \citet{dubois-etal-2025-sampling}. The ModernBERT variant we used as a supervised classifier was chosen because it is open-source and ranked high on the RAID leaderboard (available \url{https://raid-bench.xyz/leaderboard}), a popular benchmark for AI-generated text detection released at COLING 2025 and regularly updated since.

\subsection{Baselines’ results on rewritings}

\begin{figure}[htbp]
    \centering
    
    % --- Top Left Image ---
    \begin{subfigure}[b]{0.45\textwidth}
        \centering
        \includegraphics[width=\textwidth]{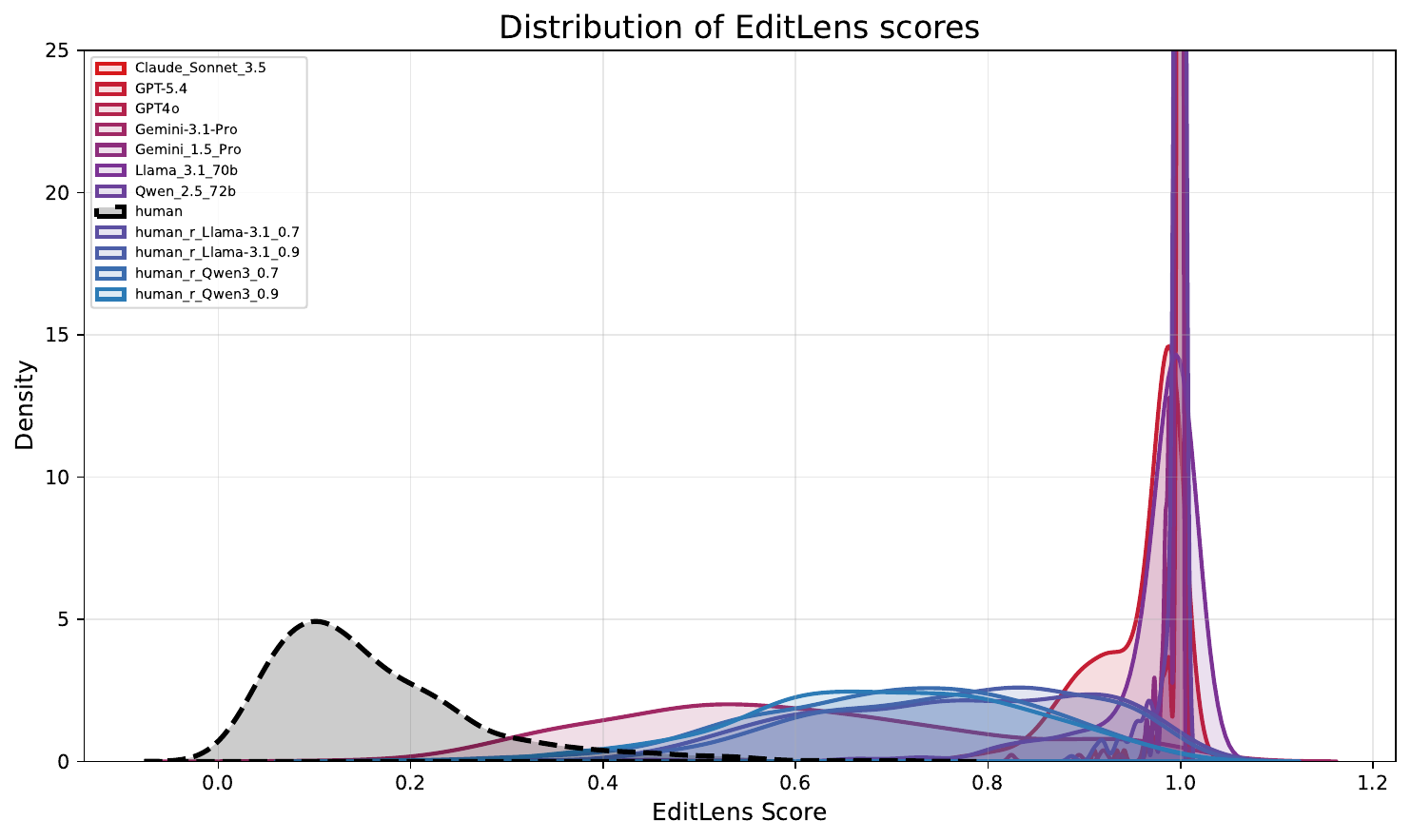} 
        \caption{Density of EditLens scores.}
        \label{fig:top_left}
    \end{subfigure}
    \hfill % Adds flexible space between the two top images
    % --- Top Right Image ---
    \begin{subfigure}[b]{0.45\textwidth}
        \centering
        \includegraphics[width=\textwidth]{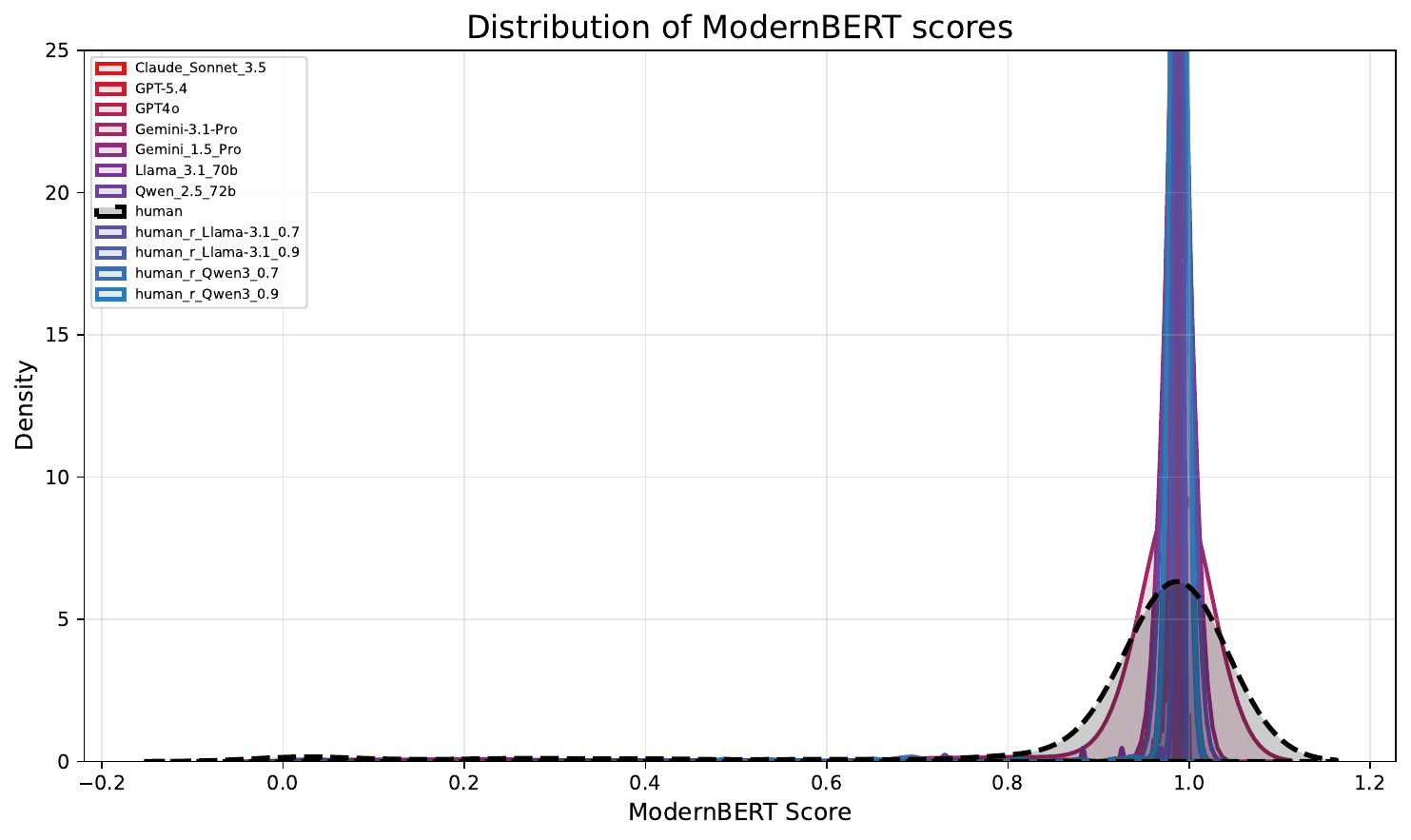} 
        \caption{Density of ModernBERT scores.}
        \label{fig:top_right}
    \end{subfigure}
    
    \vspace{1em} % Adds vertical space between the top row and bottom row
    
    % --- Bottom Left Image ---
    \begin{subfigure}[b]{0.45\textwidth}
        \centering
        \includegraphics[width=\textwidth]{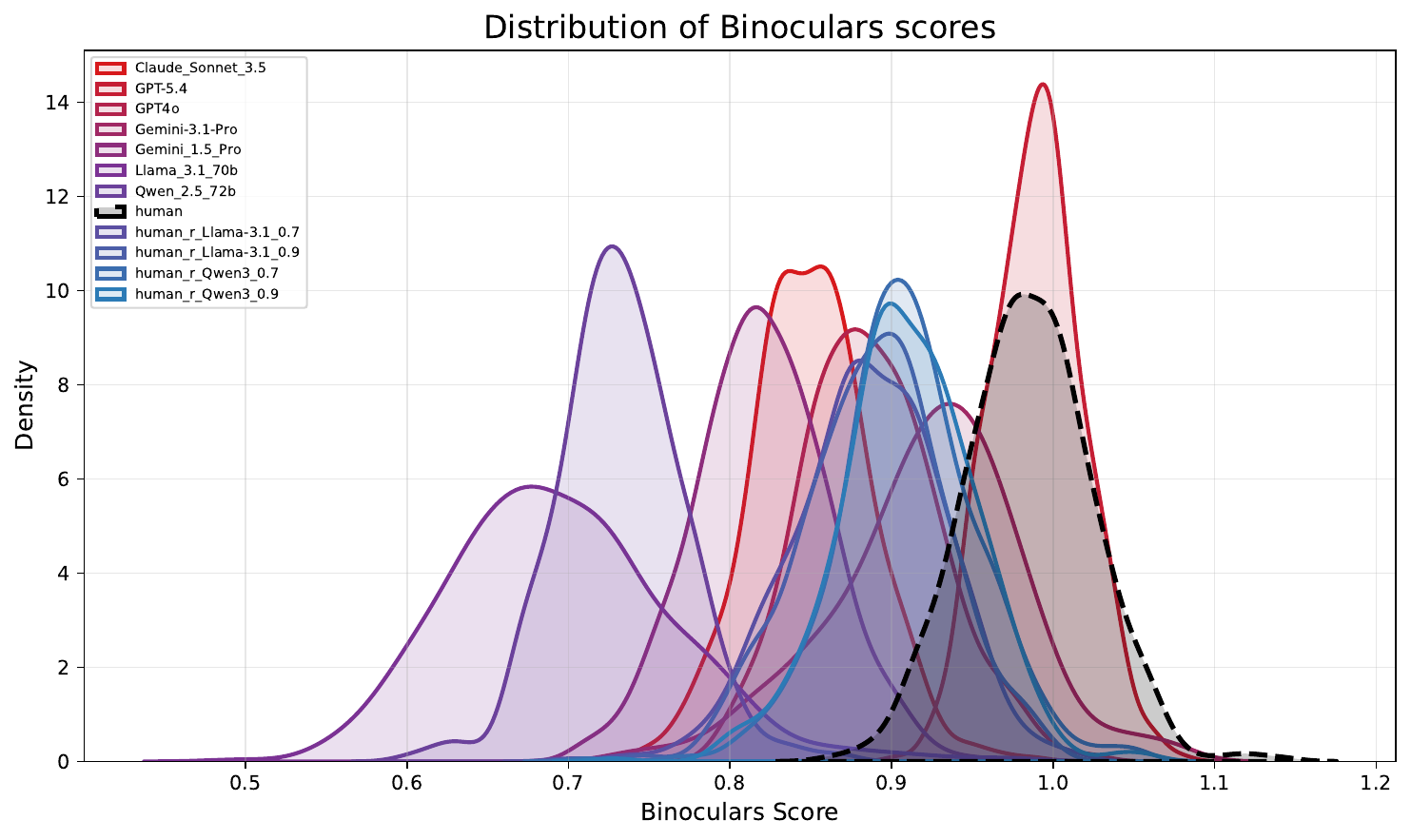} 
        \caption{Density of Binoculars scores.}
        \label{fig:bottom_left}
    \end{subfigure}
    \hfill % Adds flexible space between the two bottom images
    % --- Bottom Right Image (NEW) ---
    \begin{subfigure}[b]{0.45\textwidth}
        \centering
        \includegraphics[width=\textwidth]{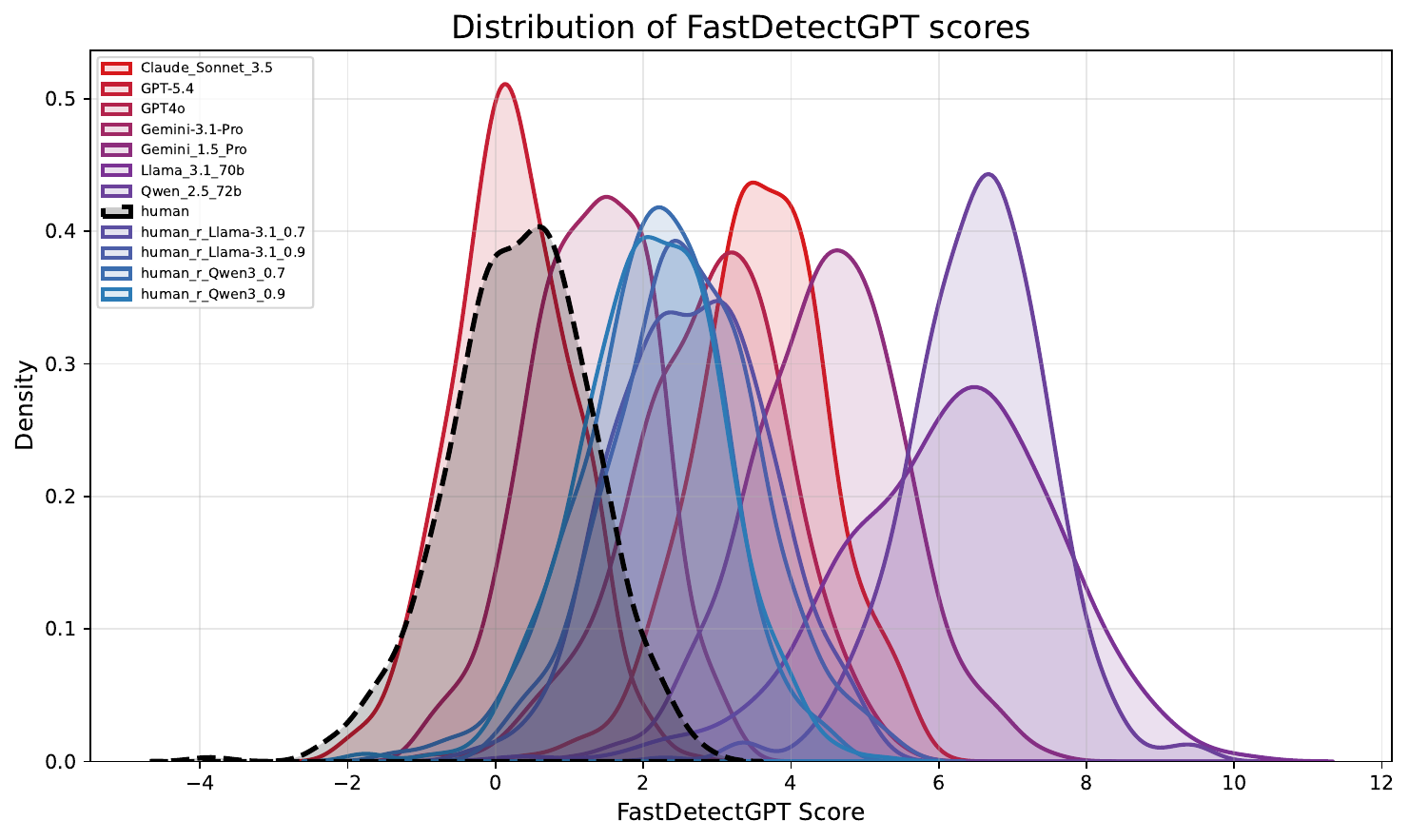} 
        \caption{Density of Fast-DetectGPT scores.} 
        \label{fig:bottom_right}
    \end{subfigure}
    
    \caption{Density of Scores for traditional AI detection methods on fully-generated reviews and polished versions of human ones. \qt{human\_r\_\{model\_name\}} labels rewritings of the human reviews by LLM \{model\_name\}.}
    \label{fig:detectors_vs_rewritings}
\end{figure}

On Figure~\ref{fig:detectors_vs_rewritings}, we can observe that EditLens tends to assign \qt{AI-editing} scores between $0.4$ and $1$ to the polished versions, corresponding to what it was trained for, estimating how much those polished versions differ from their originals. The unsupervised detectors tend to confound our polished productions with generations from Claude-Sonnet-3.5 and GPT-4o. The detector based on ModernBERT shows once again poor out-of-domain capabilities. 

%% >>> SKELETON NOTE: IMPORTED FROM B: TPR at fixed FPR, referenced from the separability section.
\FloatBarrier
\section{More Results for \method} \label{app:tpr_details}

Table~\ref{tab:our_method_perf} reports additional results for the True Positive Rate (TPR) at a fixed False Positive Rate (FPR). This is where the limits of our small detector model start to show, as illustrated by the low TPR@0.1\%FPR for some recent \emph{generator} models.

\begin{table}[htbp]
    \centering
    \caption{Results of our method across different generator models in the ICLR 2019 subset (N=500 per row). Detector model is \scoringmodel. The \qt{Machine-polished} texts are the ones mentioned in the bottom four rows of Table~\ref{tab:reviewer_metrics}, from human reviews polished by Llama-3.1-8B-IT and Qwen3-8B with temperatures $0.7$ and $0.9$.}
    \resizebox{\columnwidth}{!}{%
    \begin{tabular}{lcccc}
        & \multicolumn{4}{c}{\textbf {TPR@ ...} } \\
        \textbf{Sample Generator} & \textbf{0.1\%FPR} & \textbf{0.5\%FPR} & \textbf{1\%FPR} & \textbf{5\%FPR}   \\
        \midrule
        \multicolumn{5}{l}{\textit{\textbf{Fully-delegated vs. Human}}} \\
        \quad Claude-Sonnet-3.5   & 74.8 & 93.4 & 96.2 & 99.2 \\
        \quad GPT-4o              & 24.4 & 51.6 & 71.0 & 92.4 \\
        \quad Gemini-1.5-Pro      & 91.4 & 98.0 & 99.4 & 100.0 \\
        \quad Llama-3.1-70B       & 99.2 & 99.4 & 99.8 & 99.8 \\
        \quad Qwen2.5-72B         & 99.2 & 99.6 & 99.6 & 100.0 \\
        \quad GPT-5.4             & 0.6  & 7.2  & 18.0 & 67.6 \\
        \quad Gemini-3.1-Pro      & 15.8 & 32.2 & 46.0 & 75.2 \\
        \midrule
        \multicolumn{5}{l}{\textit{\textbf{Fully-delegated vs. Machine-polished}}} \\
        \quad Claude-Sonnet-3.5   & 79.8 & 93.8 & 95.8 & 99.2 \\
        \quad GPT-4o              & 29.6 & 53.2 & 65.2 & 90.6 \\
        \quad Gemini-1.5-Pro      & 93.4 & 98.2 & 99.4 & 100.0 \\
        \quad Llama-3.1-70B       & 99.2 & 99.4 & 99.8 & 99.8 \\
        \quad Qwen2.5-72B         & 99.4 & 99.6 & 99.6 & 100.0 \\
        \quad GPT-5.4             & 1.0  & 7.6  & 15.0 & 60.6 \\
        \quad Gemini-3.1-Pro      & 17.8 & 33.6 & 42.6 & 72.4 \\
        \bottomrule
    \end{tabular}%
    }
    \label{tab:our_method_perf}
\end{table}
\FloatBarrier
\section{Additional results in our graded experiments} \label{app:additional_results}

\subsection{Random-ordered bullet points}
\label{app:random-order-bullets}

\begin{figure}[htb]
    \centering
    \includegraphics[width=\linewidth]{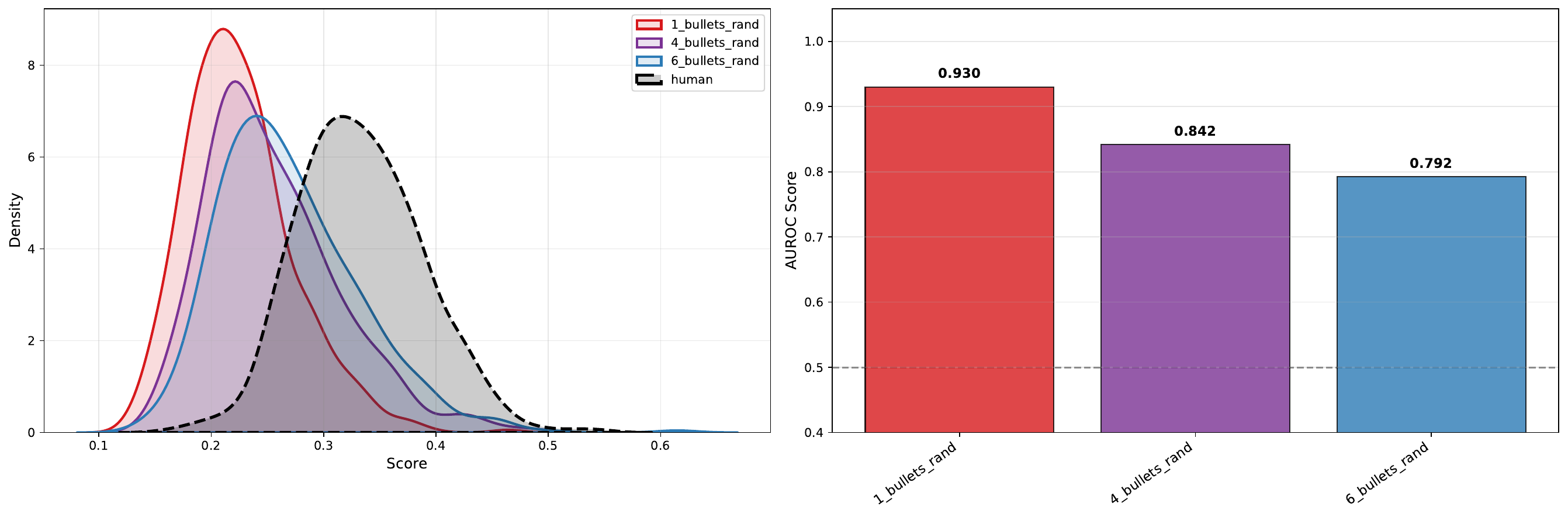}
    \caption{Reviews generated from a varying number of bullet points supplied in the prompt, sampled in
    a random order. Left: distribution of scores; right: AUC with respect to human texts.}
    \label{fig:qwen_bullets_random}
\end{figure}

Figure~\ref{fig:qwen_bullets_random} reports the control described in
Section~\ref{ssec:graded_effort}, in which the $k$ supplied bullet points
are sampled uniformly rather than taken in a fixed order. The same
decreasing AUC trend is preserved.

\subsection{Other Generators for Hybrid Reviews}
\label{app:more_bullet_points}

\begin{figure}[htbp]
    \centering
    \includegraphics[width=\linewidth]{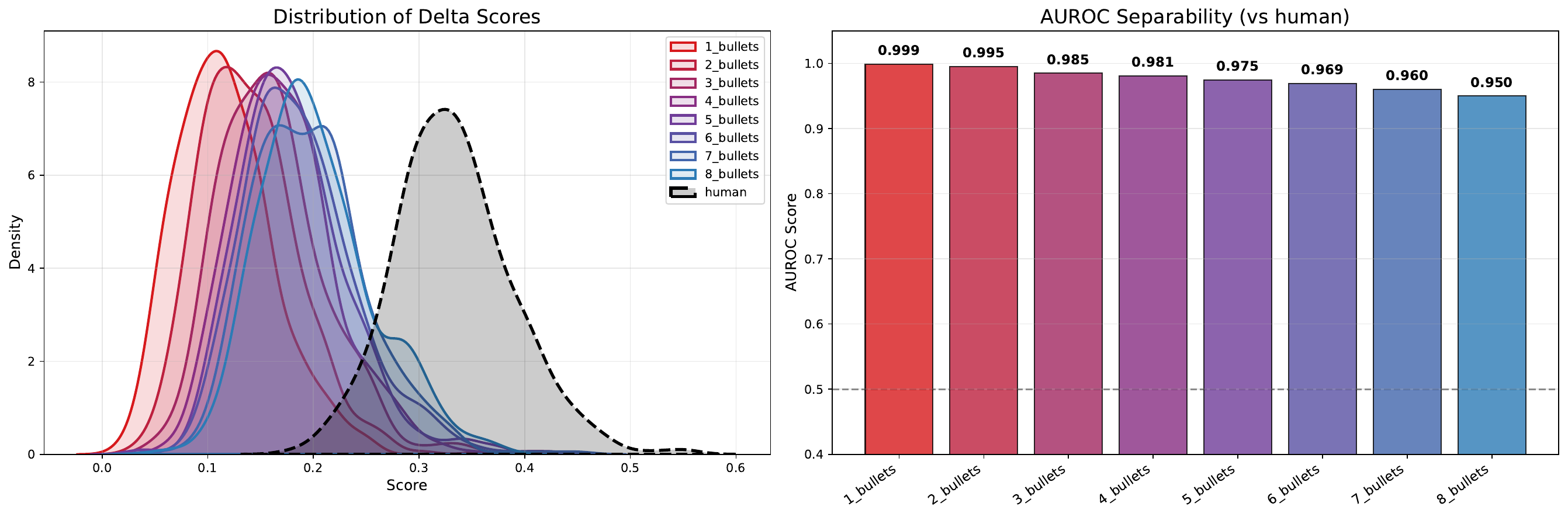}
    \caption{Reviews generated with Mistral-Small-3.2-24B-Instruct-2506 from varying amount of information given in the prompt as bullet points. On the left are the density of scores and on the right are the AUC compared to human scores.}
    \label{fig:mistral_bullets}
\end{figure}

\begin{figure}[htbp]
    \centering
    \includegraphics[width=\linewidth]{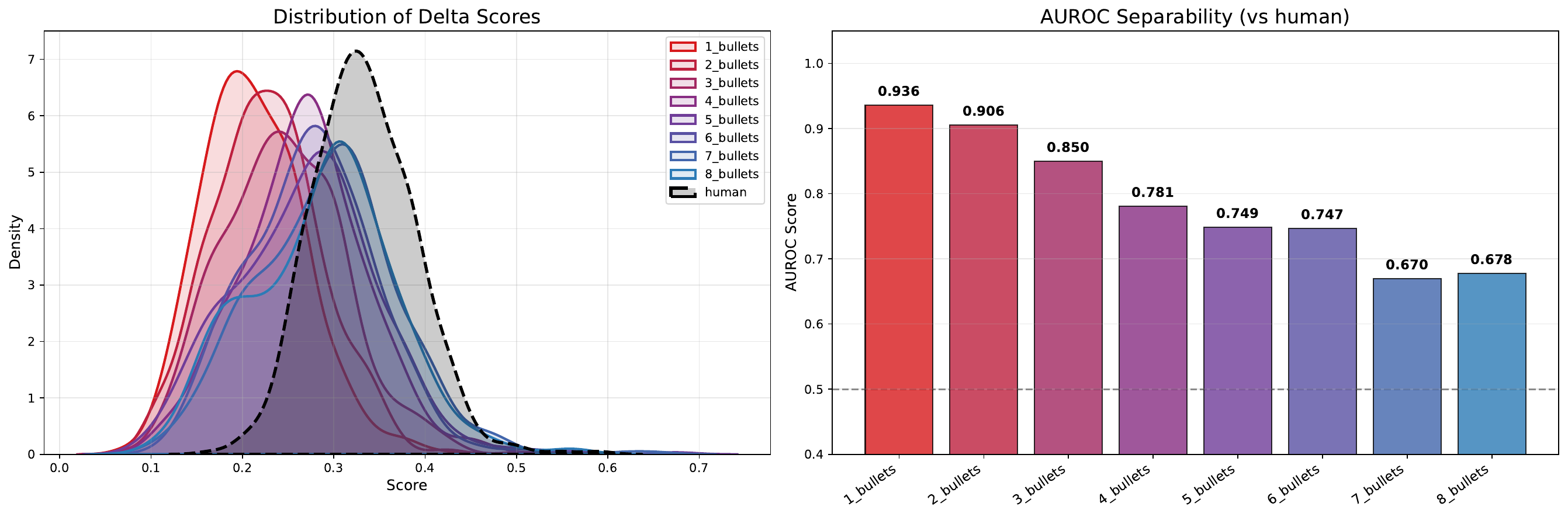}
    \caption{Reviews generated with Gemma-4-E4B-it from varying amount of information given in the prompt as bullet points. On the left are the density of scores and on the right are the AUC compared to human scores.}
    \label{fig:gemma_bullets}
\end{figure}

The reviews generated from bullet points by other models follow the same trends as the ones displayed in Figure~\ref{fig:qwen_bullets}, the more external contribution is given in the prompt, the closer to the human distribution the scores become. For Mistral (Figure~\ref{fig:mistral_bullets}), despite getting closer the score distribution remains far from the human ones, while the Gemma generations (Figure~\ref{fig:gemma_bullets}) go from far to very close.

\subsection{Other Scoring Models in Peer-review Settings}
\label{app:other_scoring_models}

%% >>> SKELETON NOTE: REMOVED: the four scoring-model figures (base paper lines 557-583), replaced by Table~\ref{tab:other_scoring_perf} in the ablations section. Label kept so that the cross-reference in Section~\ref{sec:experiment_details} still resolves.

The corresponding AUC values are reported in Table~\ref{tab:other_scoring_perf}.

\FloatBarrier
\section{NeurIPS \& Other Years} \label{app:more_data}
Going beyond ICLR 2019 data only, we also generated rewritings and ran our \method scoring method on subsets of the IntelLabs-AI-Peer-Review dataset from ICLR 2020, ICLR 2021 and NeurIPS 2021 (NeurIPS began using OpenReview in 2021). Since NeurIPS only publicly publishes \emph{accepted paper}’s reviews, there could be a bias towards positive ones, yet that does not seem to affect our score’s distribution.

\begin{figure}[!htbp]
    \centering
    \includegraphics[width=\linewidth]{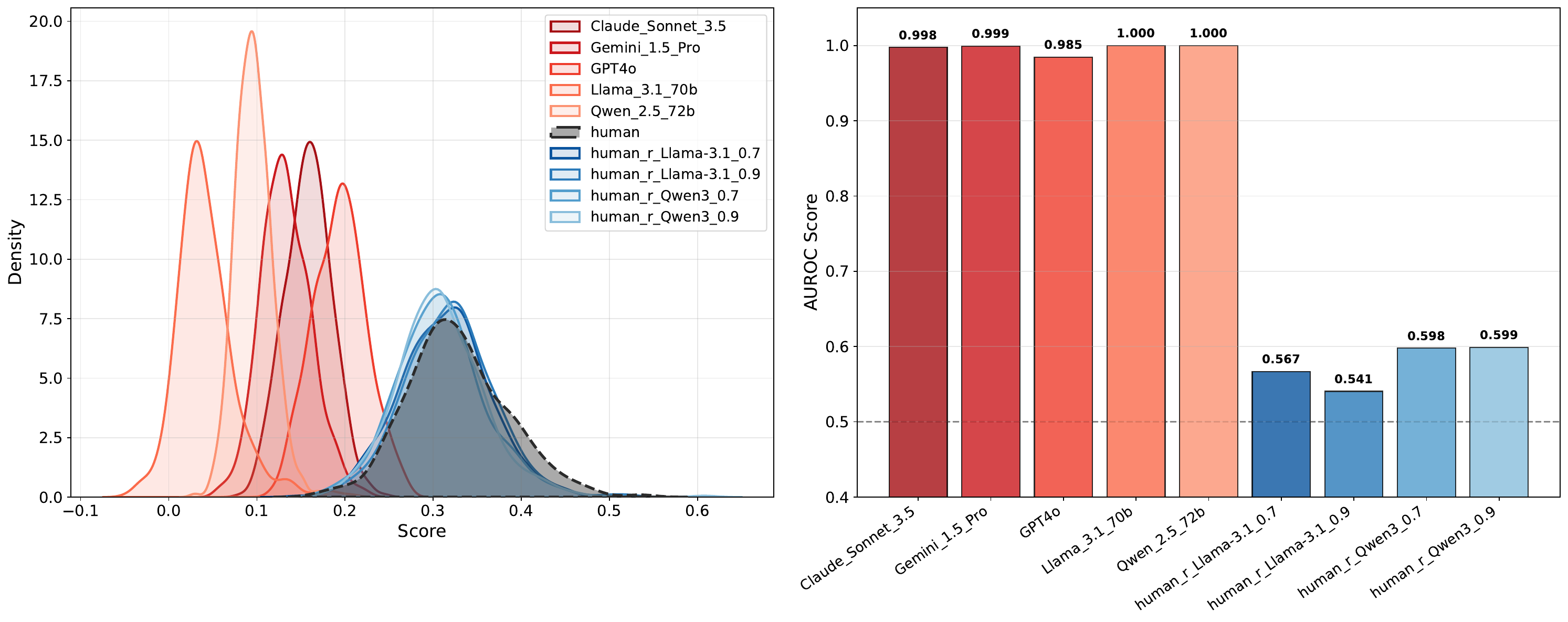}
    \caption{Distribution of \method scores for the ICLR 2020 subset of the IntelLabs-AI Peer Review dataset. (AUC values are displayed next to each model name)}
    \label{fig:iclr_2020_scores}
\end{figure}

\begin{figure}[!htbp]
    \centering
    \includegraphics[width=\linewidth]{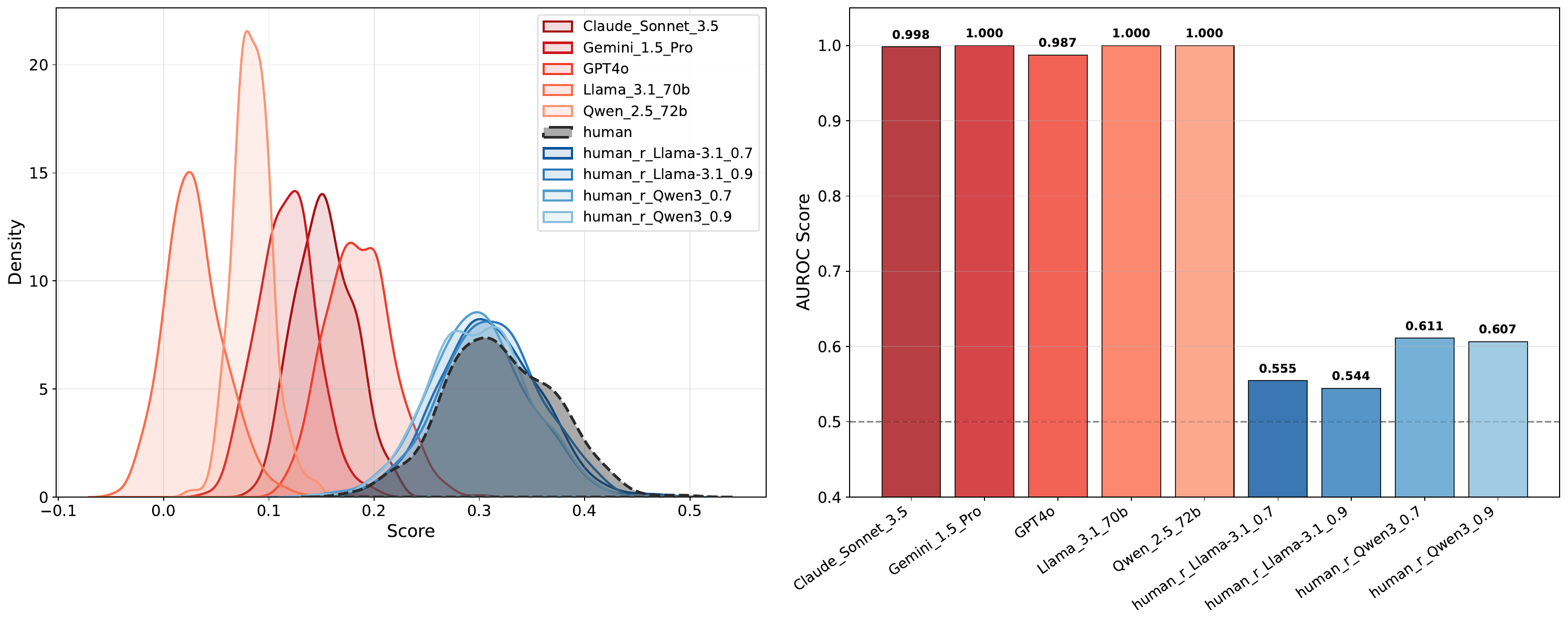}
    \caption{Distribution of \method scores for the ICLR 2021 subset of the IntelLabs-AI Peer Review dataset. (AUC values are displayed next to each model name)}
    \label{fig:iclr_2021_scores}
\end{figure}

\begin{figure}[!htbp]
    \centering
    \includegraphics[width=\linewidth]{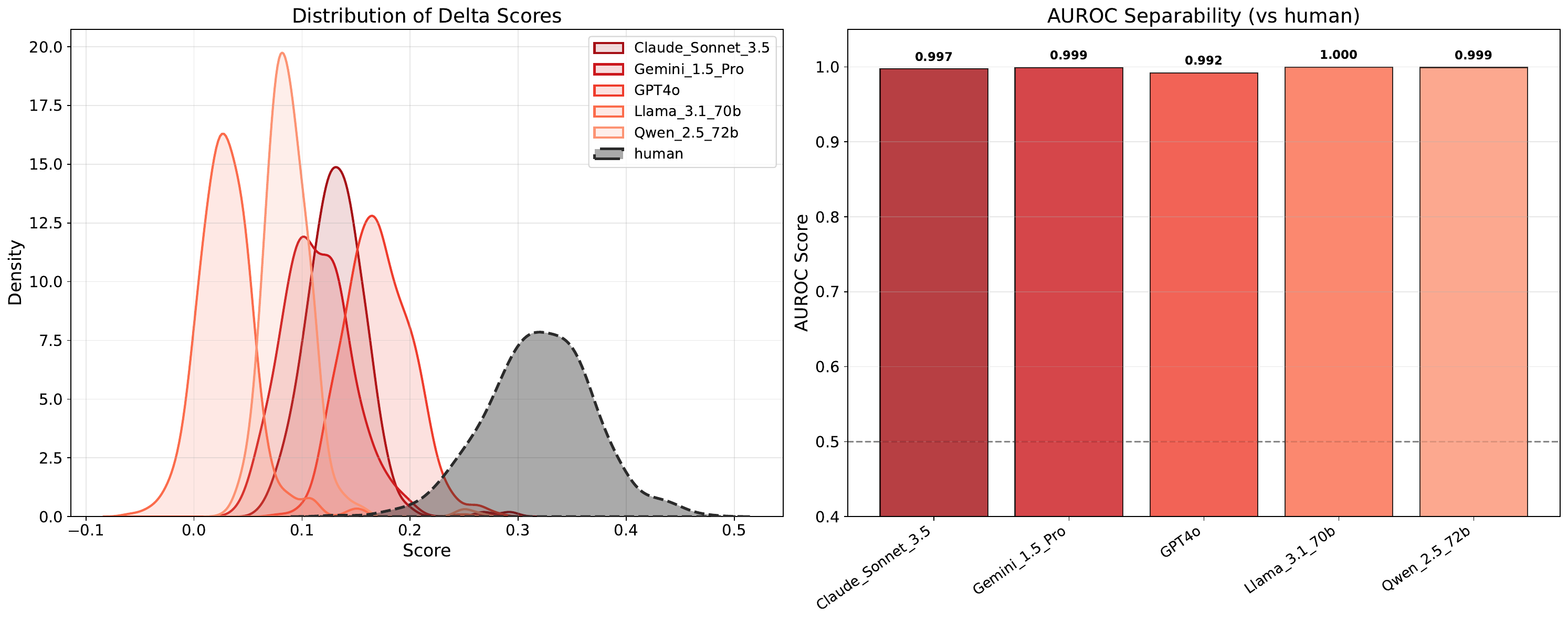}
    \caption{Distribution of \method scores for the Neurips 2021 subset of the IntelLabs-AI Peer Review dataset. (AUC values are displayed next to each model name)}
    \label{fig:neurips_2021_scores}
\end{figure}

%% >>> SKELETON NOTE: IMPORTED FROM B: high-temperature review samples, referenced from the edge-cases section.
\FloatBarrier
\section{High Temperatures Reviews} \label{app:high-temp}

\begin{lstlisting}[title={human review}]
"text": "This paper is a direct follow-up on the Sun-Qu-Wright non-convex optimization view on the Spielman-Wang-Wright complete dictionary learning approach. In the latter paper the idea is to simply realize that with Y=AX, X being nxm sparse and A a nxn rotation, one has the property that for m large enough, the rows of X will be the sparsest element of the subspace in R^m generated by the rows of Y. This leads to a natural non-convex optimization problem, whose local optimum are hopefully the rows of X. This was proved in SWW for *very* sparse X, and then later improved in SQW to the linear sparsity scenario. The present paper refines this approach, and obtain slightly better sample complexity by studying the most natural non-convex problem (ell_1 regularization on the sphere).\n\n\nI am not an expert on SQW so it is hard to evaluate how difficult it was to extend their approach to the non-smooth case (which seems to be the main issue with ell_1 regularization compared to the surrogate loss of SQW).\n\n\nOverall I think this is a solid theoretical contribution, at least from the point of view of non-smooth non-convex optimization. I have some concerns about the model itself. Indeed *complete* dictionary learning seemed like an important first step in 2012 towards more general and realistic scenario. It is unclear to this reviewer whether the insights gained for this complete scenario are actually useful more generally.\n"
\end{lstlisting}

\begin{lstlisting}[title={Llama-70 temperature=1 review}]
"text": "This paper presents a theoretical guarantee for orthogonal dictionary learning using subgradient descent on a natural \u2113_1 minimization formulation. The authors develop tools for analyzing the optimization landscape of nonconvex nonsmooth functions, which could be of broader interest. The proposed algorithm, Riemannian subgradient descent, is shown to recover the dictionary with high probability under mild statistical assumptions on the data.\n\nClarity: 9/10\nThe paper is well-written, and the authors have made a significant effort to make the content accessible to a broad audience. The notation is clear, and the mathematical derivations are well-explained. However, some of the proofs are relegated to the appendices, which might make it difficult for readers to follow the flow of the main argument.\n\nQuality: 9.5/10\nThe technical quality of the paper is excellent. The authors have developed a novel framework for analyzing the optimization landscape of nonconvex nonsmooth functions, which is a significant contribution to the field. The theoretical results are well-supported by simulations, and the experimental design is thorough.\n\nNovelty: 9/10\nThe paper presents a new approach to dictionary learning, which is a significant departure from existing methods. The use of Riemannian subgradient descent on a nonconvex nonsmooth formulation is novel and well-motivated. However, some of the ideas and techniques used in the paper, such as the use of covering numbers and the vector angle inequality, are not entirely new.\n\nSignificance: 9.5/10\nThe paper makes a significant contribution to the field of dictionary learning and nonconvex optimization. The proposed algorithm has the potential to be widely applicable, and the theoretical framework developed in the paper could be used to analyze other nonconvex nonsmooth problems. The experimental results demonstrate the effectiveness of the algorithm in practice.\n\nMinor suggestions:\n\n* The authors could consider providing more intuition behind the choice of the \u2113_1 minimization formulation and the use of Riemannian subgradient descent.\n* Some of the proofs in the appendices could be shortened or omitted, and the main argument could be made more self-contained.\n* The experimental section could be expanded to include more comparisons with existing methods.\n\nMajor suggestions:\n\n* The authors could consider generalizing their results to overcomplete dictionaries, which is an important open problem in the field.\n* The paper could benefit from more discussion on the implications of the theoretical results for practice, and how the algorithm could be used in real-world applications.\n\nOverall, this is a strong paper that makes significant contributions to the field of dictionary learning and nonconvex optimization. With some minor revisions, it has the potential to be a top-tier paper in the conference.",
\end{lstlisting}

\begin{lstlisting}[title={Llama-70 temperature=1.5 review}]
    "text": "This paper proposes a subgradient descent algorithm to solve the non-convex, non-smooth dictionary learning problem. The authors demonstrate that their method is guaranteed to recover orthogonal dictionaries in polynomial time, under a statistical model and a natural \u21131 minimization formulation. This is an improvement over existing methods that require expensive computations or initialization schemes. The paper also explores the use of nonsmooth optimization on Riemannian manifolds and develop tools for analyzing nonconvex, nonsmooth optimization problems.\n\nStrengths:\n1. The paper tackles a challenging non-convex, non-smooth problem, providing a solution with both theoretical guarantees and strong experiments.\n2. The use of Riemannian manifold optimization and Clarke subdifferentials for locally Lipschitz functions is innovative and shows promising results.\n3. The paper contributes valuable insights into the landscape of the dictionary learning objective and the behavior of subgradient descent.\n4. The comparison with existing methods, highlighting the advantages of their approach in terms of computational efficiency and simplicity, is instructive.\n5. Synthetic and real-image experiments demonstrate the effectiveness of their method, complemented by detailed and thorough analysis.\n\nWeaknesses:\n1. While a nice result is established for orthogonal dictionaries, the extension to overcomplete dictionaries, as discussed in the paper, remains to be studied.\n2. Although sample complexity is claimed to be low, in fact, their best analysis indicates an n^4 term; future research is necessary to improve these upper bounds.\n3. There remains a challenge in handling over-complete dictionaries with the tools currently at their disposal.\n\n\n\nSuggestions for Improvement and Research Directions:\nWhile this is a substantial body of work by the authors. Address and study how these insights for nonsmooth optimization might be generalized for the problem beyond this paper. Future lines of research could:\n- Investigate how the insights for nonsmooth optimization for dictionary learning can be extended to deep learning. Specifically explore application on ReLU networks.\n\nClarity:\n9\nQuality:\n9\nNovelty:\n9\nSignificance:\n8",
\end{lstlisting}

\begin{lstlisting}[title={Llama-70 temperature=2.0 review}]
"text": "The paper proposes a subgradient descent algorithm for orthogonal dictionary learning, recovering dictionaries through a sparse coding problem by solving an \u21131-norm minimization formulation. The work provides a rigorous theoretical analysis, characterizing landscapes of nonsmooth, nonconvex functions, particularly for learning compact representations.\n\nKey Contributions and Merits:\n(1) Rigorous theoretical guarantee for orthogonal dictionary learning with subgradient descent, without elaborate initialization schemes, thus offering a first result in solving the task in terms of the \"sparsity-in-spacity\"-paragidem of Donho  et al\n    .  \n\n( 2   )\nAn analytical landscape, characterise which allows simple gradient descent in learning dictionaries via random initializaton and a scalable algorithm to solve  compact sparity, despite lack of  lipschiz and differnciabil- ity and in presence of large dictionary.\n\nMinor issues of Technical writing :\nIn an effort  notto misundersood this contribu-ions - we make some suggestions below, for improvement for final text.\n\n   .There should be clear and concise headings to identify intro background theoretrcal results algoritmis simulaton experiments.(minor suggestion : one  may break in many sections such background - may help better organisation.\n\n    One sentence which goes like  one might say \"the paper provide rigeros theory on orthognol diractonal\", sounds a bit non technical phrase one is referri-  ng  can improve clarity, like refer \" provides rigorous landscape analysis\".  \n      \nThe experimental result should give some clear interpretation - of empirical study - in term of figure  table and references.\n\n Clarity is 9 out of 10 point.\n\nSignificance This papa make advance results for a task learning with sparset dictionary which  widely applicble real word applivcation.\n\nnoveltity- paper combines of new insights tools  results of analysis from optim- nonsomouth theory to an appliclation (dictionary)\n\nReccomendation of Publication :Yes",
\end{lstlisting}

\newpage
\section{Prompt details} \label{app:prompts}
\subsection{Peer Review Prompts}

In the context of Peer-Reviews, the context $C$ in Equation~\ref{eq:score-average} is made-up of the paper’s text along with a generic prompt $g$, detailed below:
\begin{lstlisting}
    "You are an expert reviewer for the ICLR Conference. "
    "Your task is to write a high-quality peer review. "
    "Critique based on: Clarity, Quality, Novelty, and Significance."
\end{lstlisting}

For fully generated peer reviews, we followed the prompting setup of \citet{yu-etal-2026-is-your-paper} when querying GPT-5.4 and Gemini-3.1-Pro. Each model was given the same system and user prompts, together with the conference reviewer guidelines, the review template, and the human decision, so as to produce reviews as close as possible to the human ones. Because our experiments target ICLR 2019, we used the reviewer guidelines from \url{https://iclr.cc/Conferences/2019/Reviewer_Guidelines}
. GPT-5.4 was queried through the OpenAI API in March 2026 with temperature $0.9$ and \verb|max_completion_tokens|$=2048$, using the identifier \verb|gpt-5.4|. Gemini-3.1-Pro was queried during the same period through the \verb|google.genai| API, using \verb|gemini-3.1-pro-preview| and the same generation parameters.

\begin{lstlisting}
SYSTEM_PROMPT = You are an AI researcher reviewing a paper submitted to a prestigious AI research conference.
You will be provided with the manuscript text, the conference's reviewer guidelines, and the decision for the paper.
Your objective is to thoroughly evaluate the paper, adhering to the provided guidelines, and return a detailed assessment that supports the given decision using the specified response template.
Ensure your evaluation is objective, comprehensive, and aligned with the conference standards.
{reviewer_guideline}
{review_template}
\end{lstlisting}

\begin{lstlisting}
USER_PROMPT = Here is the paper you are asked to review. Write a well-justified review of this paper that aligns with a '{human_reviewer_decision}' decision: 
{text}

\end{lstlisting}

\begin{lstlisting}
REVIEWER_GUIDELINE = Publication standards. It's important to maintain the high quality of ICLR papers. At the same time, we must be cognizant of the fact that different reviewers, authors, and even different areas will have different standards about what constitutes a high-quality paper. Reviewers are encouraged to use their best judgement, but also ask themselves the following questions:

    Does the paper present substantively new ideas or explore an underexplored or highly novel question? Papers that take risks and study a less explored area are likely to have less polished results, papers that study a highly explored topic are likely to have more polished results. This phenomenon often results in reviewers excessively penalizing papers that explore underexplored topics, and it is worth accounting for this.

    At the same time, papers that present state-of-the-art results on well-studied problems should be given consideration: addressing problems that are clearly of interest to a large segment of the community is important, when the results substantively advance the state of the art.

    Ask yourself: will a substantial fraction of the ICLR attendees be interested in reading this paper?

    Another question to ask yourself: would I send this paper to one of my colleagues to read? This is often a good indication that the paper is worth accepting, though not the only indication.

At the same time, it is critical to maintain a high standard in terms of scientific rigor: if you believe that a paper has flaws in terms of its evaluation or validation, proofs, or other parts of the discussion, it is critical to point this out to the authors. The authors will have an opportunity to address these concerns, and the iterative process of improving papers after reviewer feedback is important for ensuring the highest quality of ICLR papers.

Topic fit. The scope of ICLR has increased over the years, and covers a large segment of machine learning research and applications. If you are uncertain about whether a paper is a good topic fit for ICLR, feel free to contact your area chair.

Anonymity. ICLR is double-blind, which means that authors are not aware of reviewer identities and reviewers are not aware of author identities. If you believe a paper contains an anonymity violation, contact your AC immediately. Anonymity violations are not considered as part of reviewing criteria, they are requirements for submission. Unless your AC decides that the paper does indeed violate anonymity, proceed to review it as normal. Do not reveal your or the authors' identities in the discussion.

Length. The recommended paper length is 8 pages, with unlimited additional pages for citations.  There is a strict upper limit of 10 pages.   Authors may use as many pages of appendices (after the bibliography) as they wish. Reviewers are not required to read all appendices. However, details that are necessary to understand whether a paper is reproducible or to validate proofs may be found in these appendices. This is acceptable, and it is up to the reviewers to verify they are present and correct if they have concerns about reproducibility and correctness. Papers may exceed 8 pages in the main text, but reviewers should encourage authors to avoid overly lengthy papers. In some cases, longer papers may be acceptable. For example, a paper with two pages of figures (e.g., image or video generation) might have a total of 10 pages in the main body. But the overall time to read a paper should be comparable to that of a typical 8-page conference paper. Reviewers may apply a higher reviewing standard to papers that substantially exceed this length. If reviewers believe that a paper is too lengthy, first it should be pointed out to the authors to see if they can fix it during the revisions phase.

Similar/concurrent submissions. ICLR often sees multiple concurrent submissions on similar topics from different authors. In such cases, special care should be taken to ensure that concurrent papers are reviewed to the same standard. If you are reviewing a paper and believe there is another ICLR submission that is concurrent and similar, please contact your AC to let them know.

Arxiv and prior work. ICLR considers unpublished arxiv papers to be prior work. While we encourage reviewers to apply the reasonable standards of the relevant community in considering what does and does constitute prior work, the following minimum standards will be enforced: no paper will be considered prior work if it appeared on arxiv, or another online venue, less than 30 days prior to the ICLR deadline.
\end{lstlisting}

\begin{lstlisting}
REVIEW_TEMPLATE = ""
#There was no review template given in 2019
\end{lstlisting}

To obtain the rewritten peer-reviews (human reviews then polished by AI), we prompted a Llama and a Qwen model with \verb|max_new_tokens| at 2048, and temperatures of $0.7$ and $0.9$ :
\begin{lstlisting} 
    REWRITING_PROMPT: "You are a professional scientific editor."
    "Your task is to rewrite the following peer review to improve its clarity, flow, and grammar. "
    "Do NOT change the technical content, the sentiment (positive/negative), or the scores. "
    "Keep the critique exactly as it is, just make it easier to read.\n\n"
    f"--- ORIGINAL REVIEW ---\n{original_text}\n--- END REVIEW ---\n\n"
    "Provide the rewritten version below."
\end{lstlisting}

To obtain the Bullet-points summaries of human reviews, we employed Llama-3.1-8B-Instruct in a greedy decoding mode, along with the following prompt:

\begin{lstlisting}
    BULLET_INSTRUCTION = """You are helping build a controlled benchmark for human-effort analysis.

Summarize the peer review below into exactly 8 concise bullet points, in this exact order:
1. Paper contribution
2. Main strengths
3. Main weaknesses
4. Evidence quality
5. Novelty
6. Clarity
7. Missing experiments
8. Final judgment

Requirements:
- Return exactly 8 bullet points.
- Keep each bullet concise and faithful to the review.
- Do not invent content not supported by the review.
- If a category is not clearly discussed in the review, say so briefly.
- Do not add any introduction or conclusion.
- Use the following exact format:

- Paper contribution: ...
- Main strengths: ...
- Main weaknesses: ...
- Evidence quality: ...
- Novelty: ...
- Clarity: ...
- Missing experiments: ...
- Final judgment: ...

Peer review:
\"\"\"{review_text}\"\"\"
"""
\end{lstlisting}

And then, utilising those summaries, we prompted models to generate reviews from partial human information, trying to follow a template similar to the one used for fully-machine reviews:

\begin{lstlisting}
    REVIEWER_GUIDELINE = """
Publication standards. It's important to maintain the high quality of ICLR papers. At the same time, we must be cognizant of the fact that different reviewers, authors, and even different areas will have different standards about what constitutes a high-quality paper. Reviewers are encouraged to use their best judgement, but also ask themselves the following questions:
    - Does the paper present substantively new ideas or explore an underexplored or highly novel question? 
    - Will a substantial fraction of the attendees be interested in reading this paper?
    - Would I send this paper to one of my colleagues to read? 

At the same time, it is critical to maintain a high standard in terms of scientific rigor: if you believe that a paper has flaws in terms of its evaluation or validation, proofs, or other parts of the discussion, it is critical to point this out to the authors.
"""

SYSTEM_PROMPT = f"""You are an AI researcher reviewing a paper submitted to a prestigious AI research conference.
You will be provided with the manuscript text, the conference's reviewer guidelines, and a partial list of core arguments/bullet points extracted from a previous human review.

Your objective is to thoroughly evaluate the paper, adhering to the provided guidelines.
CRITICAL INSTRUCTIONS:
1. You must use the provided bullet points as the absolute foundation of your review. 
2. Expand upon these specific points using the manuscript text to ground your arguments in reality.
3. Your final assessment MUST perfectly reflect the exact sentiments of the provided bullet points. Do not invent major critiques or praises that contradict them.
4. Even though you may only receive a few bullet points, you must write a complete, cohesive peer review.
5. DO NOT output any internal thinking process, reasoning, or preamble. Output ONLY the final review text.

Ensure your evaluation is objective, comprehensive, and aligned with the conference standards.
{REVIEWER_GUIDELINE}"""

USER_PROMPT = """Target Review Length: Approximately {target_word_count} words.

Here are the core arguments you MUST include and expand upon in your review:
---
{partial_bullets}
---

Here is the paper you are asked to review. Write a well-justified review of this paper that incorporates the core arguments above. 
OUTPUT ONLY THE REVIEW TEXT. NO PREAMBLE. NO THINKING PROCESS.
---
{text}
---"""
\end{lstlisting}

% \subsection{Essays}
% The prompts used to generate the essays and their rewritten versions are much simpler.

% \begin{lstlisting}
% persuasive_essay="Act as a middle school student, please read the below prompt and write an essay: {}"
% source_essay="Act as a middle school student, please read the source essay and write an essay based on given topic prompt. {}"
% polish="Polish and optimize the following essay: {}"
% \end{lstlisting}

% The context $C$ is similarly built depending if the essay is persuasive of based on a source document.

% \begin{lstlisting}
%             if essay_type == "persuasive":
%             context_text = f"Act as a middle school student, please read the below prompt and write an essay: {prompt_text}"
%         elif essay_type == "source-dependent":
%             context_text = f"Act as a middle school student, please read the source essay and write an essay based on given topic prompt. {prompt_text}"
% \end{lstlisting}

% \subsection{Reference Letters}
% For the Reference letters, we did not generate anything so there is no generation prompt. However, the context $C$ is built to resemble the prompts used by the original authors. 

% \begin{lstlisting}
%     context_text = f"Please write a reference letter for {name}, an {occupation}."
% \end{lstlisting}

\subsection{Graded Rewriting Prompts}

\begin{lstlisting}
            "light": "Light modification: Fix only obvious typos and grammatical errors. Strictly preserve the original vocabulary, voice, and sentence structure.",
            "moderate": "Moderate modification: Improve flow and readability. Fix awkward phrasing and smooth out transitions while maintaining the original meaning and general paragraph structure.",
            "heavy": "Heavy modification: Substantially rewrite the text for a highly professional academic tone. Reorganize sentences and upgrade vocabulary for maximum clarity and impact.",
            "major": "Major modification: Completely overhaul and paraphrase the text. Transform the review into a highly polished, impeccably written academic critique, prioritizing formal scientific language."
\end{lstlisting}

\FloatBarrier
\section{Further Details About Hint Selection}
\label{app:hint_selection}

All hint-selection methods start from a target review $T$ and construct an
auxiliary hint $x$, which is prepended to the baseline context in the form
\texttt{Hints: \{x\}}. The score is computed by comparing the token-level
log-probabilities of $T$ under $P_\theta(\cdot\mid C,x)$ and
$P_\theta(\cdot\mid C)$. Most methods operate in word space, while
surprisal-based spans are selected in token space. Unless otherwise
specified, $r\in(0,1]$ controls the disclosure budget.

\paragraph{Random words.}
Let $T=(w_1,\dots,w_m)$ be the sequence of words in the target text. We sample
\[
k=\left\lceil r m \right\rceil
\]
word positions uniformly at random without replacement, then reorder the selected words according to their original positions and concatenate them. This yields a sparse lexical hint with no guarantee of local coherence.

\paragraph{Prefix / suffix disclosure.}
For prefix disclosure, we reveal the first $k=\lceil r m \rceil$ words of $T$; for suffix disclosure, we reveal the last $k$ words. These methods preserve discourse order and local coherence, but are strongly dependent on the formatting and rhetorical organization of the text.

\paragraph{Random n-grams.}
The text is first partitioned into non-overlapping contiguous word chunks of fixed length $n$ (the \texttt{span\_length} parameter). We then sample enough chunks to cover approximately a fraction $r$ of the text:
\[
k_{\mathrm{chunks}}=
\left\lceil \frac{r m}{n} \right\rceil.
\]
The selected chunks are concatenated in their original order. Compared with random words, this estimator preserves more local semantic and syntactic structure.

\paragraph{Surprisal spans.}
This estimator uses the baseline token log-probabilities of the target text under $P_\theta(\cdot \mid C)$. Let $(t_1,\dots,t_n)$ denote the tokenised target and let $\ell_i=\log P_\theta(t_i \mid C,t_{<i})$ be the corresponding baseline token log-probabilities. We define token surprisals as
\[
s_i=-\ell_i.
\]
Using a sliding window of length $L$ tokens, each candidate span is scored by the sum of its surprisals:
\[
S(j)=\sum_{i=j}^{j+L-1} s_i.
\]
We then greedily select the highest-scoring non-overlapping windows until the token budget
\[
k_{\mathrm{tok}}=\left\lceil r n \right\rceil
\]
is reached. The selected token spans are decoded back to text and concatenated. This estimator can be viewed as a near-oracle self-revelation baseline, since it reveals the least predictable parts of the text relative to the context.

\paragraph{Keywords.}
Keyword extraction is based on a corpus-level TF--IDF model. We first fit a \texttt{TfidfVectorizer} on the full corpus of reviews. For a given target text $T$, we compute its TF-IDF vector and associate each normalised word form with its TF-IDF score. Each word occurrence in $T$ is then scored by the TF-IDF value of its lowercased, punctuation-stripped form. We keep the top
\[
k=\left\lceil r m \right\rceil
\]
word occurrences by score, then restore their original order and concatenate them. This produces a compact lexical hint that emphasizes content-specific terms while ignoring common review vocabulary.

\paragraph{Top-ranked sentences.}
Key sentences are also built from the corpus-level TF--IDF model. The target text is first segmented into sentences using a sentence tokenizer. If the text contains at most $n_s$ sentences, where $n_s$ is the summary budget, the whole text is returned. Otherwise, each sentence is scored by the average TF--IDF value of its normalised word forms:
\[
\mathrm{score}(\sigma)=
\frac{1}{|\sigma|}
\sum_{w \in \sigma} \mathrm{TFIDF}(w).
\]
We then select the top $n_s$ sentences by score and restore them to their original order before concatenation. This method yields a lightweight extractive sentence subset that preserves coherence better than isolated keywords while remaining fully self-supervised.

\paragraph{Remarks.}
The keyword and top-ranked sentences methods both rely on a TF--IDF model fitted on the entire corpus and therefore measure salience relative to the dataset rather than to the individual document alone. By contrast, random, prefix/suffix, and random-span methods are purely local to the target text. Surprisal spans are the only hint family that depends directly on the baseline behavior of the scoring model $P_\theta$.

\FloatBarrier
\section{Computational Cost}
\label{app:complexity}
\subsection{Theoretical Complexity}
The computational cost of our framework is dominated by repeated scoring under the proxy language model $P_\theta$. For a given example, let $C$ denote the context, $T$ the target text, and $x$ a selected hint. The core score compares
$\log P_\theta(T\mid C)$ and $\log P_\theta(T\mid C,x),$ so the basic computation requires one \emph{baseline} forward pass and one \emph{hinted} forward pass over the same target text. In the simplest setting, where a single hint is constructed deterministically, the cost is therefore equivalent to \textbf{two forward passes} of the scoring model.

More generally, our experiments average over multiple hint realizations. If $N$ hints are sampled for the same text, the total cost becomes $1 + N$ forward passes: one baseline pass to obtain $\log P_\theta(T\mid C)$, and $N$ additional passes to obtain $\log P_\theta(T\mid C,x_j)$ for each hint $x_j$. In our implementation, stochastic estimators such as random words and random n-grams use multiple repeats, while deterministic estimators such as prefix and suffix disclosure use only one. Surprisal-based spans are also deterministic once the baseline pass has been computed, so they likewise require only one additional hinted pass.

Let $L$ denote the total sequence length seen by the model during scoring, that is, the tokenised length of the context, prompt wrapper, and target text. The exact cost of one forward pass depends on the underlying architecture and attention implementation, but for a standard Transformer it scales roughly quadratically with sequence length, i.e.
\[
\mathcal{O}(L^2),
\]
with memory usage of the same order. Consequently, the end-to-end cost per example is approximately
\[
\mathcal{O}\big((1+N)L^2\big),
\]
up to implementation-dependent constants. In practice, the dominant factor is therefore the number of scoring passes and the effective sequence length, rather than the hint-selection procedure itself. An optimisation that we have not put yet into practice for stochastic calculations would be to compute the forward pass over $C$ once and then re-use that shared KV cache across all $N$ repeats of our hint selection method. 

By comparison, hint extraction is relatively cheap. Prefix, suffix, random-word, and random-span methods are linear in the length of the text. Keyword and top-ranked sentence hints require a TF-IDF transform and lightweight ranking operations, which remain negligible relative to model scoring. Surprisal-based spans are the most expensive hint family, but only because they rely on the baseline pass through $P_\theta$; the span-selection step itself is inexpensive once token log-probabilities have been obtained.

Overall, the practical bottleneck of the method is not constructing hints, but repeatedly evaluating the scoring model on long contexts. In particular, when $N>1$, stochastic hint families trade additional robustness for a near-linear increase in inference cost.

\subsection{Hardware Used}
All experiments were conducted on NVIDIA H100 GPUs with 80\,GB of memory, using bfloat16 precision. The total compute budget for the project is approximately 4000 GPU-hours, including exploratory runs, ablations, and generation experiments. This estimate excludes generations obtained through proprietary APIs, such as GPT and Gemini, since the underlying hardware and runtime are not exposed to the user. As mentioned above, most of this compute was spent on scoring passes under the proxy model.

\FloatBarrier
\section{Additional Results For All Hint Selection Methods} \label{app:mega_ablation}
% --- PEER REVIEWS ---

\begin{table}[htbp]
\centering
\caption{AUC scores for Peer Reviews generated with Qwen3.5 using the Random Words hint method.}
\label{tab:pr_random}
\begin{tabular}{l ccccc}
\toprule
Group (vs human) & r=0.1 & r=0.2 & r=0.3 & r=0.4 & r=0.5 \\
\midrule
1\_bullets & 0.809 & 0.925 & 0.956 & 0.968 & 0.977 \\
2\_bullets & 0.804 & 0.888 & 0.926 & 0.948 & 0.958 \\
3\_bullets & 0.789 & 0.860 & 0.898 & 0.924 & 0.939 \\
4\_bullets & 0.756 & 0.816 & 0.865 & 0.892 & 0.911 \\
5\_bullets & 0.737 & 0.795 & 0.839 & 0.876 & 0.896 \\
6\_bullets & 0.750 & 0.801 & 0.831 & 0.865 & 0.881 \\
7\_bullets & 0.730 & 0.776 & 0.806 & 0.836 & 0.868 \\
8\_bullets & 0.715 & 0.760 & 0.792 & 0.836 & 0.851 \\
\bottomrule
\end{tabular}
\end{table}

\begin{table}[htbp]
\centering
\caption{AUC scores for Peer Reviews generated with Qwen3.5 using the Prefix Disclosure method.}
\label{tab:pr_prefix}
\begin{tabular}{l ccccc}
\toprule
Group (vs human) & r=0.1 & r=0.2 & r=0.3 & r=0.4 & r=0.5 \\
\midrule
1\_bullets & 0.873 & 0.920 & 0.958 & 0.976 & 0.985 \\
2\_bullets & 0.871 & 0.920 & 0.944 & 0.957 & 0.965 \\
3\_bullets & 0.866 & 0.898 & 0.921 & 0.938 & 0.942 \\
4\_bullets & 0.856 & 0.886 & 0.906 & 0.916 & 0.921 \\
5\_bullets & 0.848 & 0.871 & 0.890 & 0.907 & 0.914 \\
6\_bullets & 0.853 & 0.867 & 0.889 & 0.902 & 0.904 \\
7\_bullets & 0.836 & 0.860 & 0.875 & 0.884 & 0.882 \\
8\_bullets & 0.840 & 0.854 & 0.871 & 0.884 & 0.883 \\
\bottomrule
\end{tabular}
\end{table}

\begin{table}[htbp]
\centering
\caption{AUC scores for Peer Reviews generated with Qwen3.5 using the Suffix Disclosure method.}
\label{tab:pr_suffix}
\begin{tabular}{l ccccc}
\toprule
Group (vs human) & r=0.1 & r=0.2 & r=0.3 & r=0.4 & r=0.5 \\
\midrule
1\_bullets & 0.806 & 0.879 & 0.915 & 0.946 & 0.961 \\
2\_bullets & 0.803 & 0.880 & 0.917 & 0.942 & 0.954 \\
3\_bullets & 0.773 & 0.856 & 0.894 & 0.917 & 0.929 \\
4\_bullets & 0.746 & 0.839 & 0.866 & 0.890 & 0.903 \\
5\_bullets & 0.738 & 0.842 & 0.868 & 0.880 & 0.896 \\
6\_bullets & 0.756 & 0.837 & 0.863 & 0.879 & 0.886 \\
7\_bullets & 0.747 & 0.837 & 0.858 & 0.873 & 0.881 \\
8\_bullets & 0.751 & 0.824 & 0.835 & 0.863 & 0.867 \\
\bottomrule
\end{tabular}
\end{table}

\begin{table}[htbp]
\centering
\caption{AUC scores for Peer Reviews generated with Qwen3.5 using the Keywords Extraction method.}
\label{tab:pr_keywords}
\begin{tabular}{l ccccc}
\toprule
Group (vs human) & r=0.1 & r=0.2 & r=0.3 & r=0.4 & r=0.5 \\
\midrule
1\_bullets & 0.749 & 0.817 & 0.921 & 0.963 & 0.964 \\
2\_bullets & 0.721 & 0.778 & 0.886 & 0.942 & 0.943 \\
3\_bullets & 0.652 & 0.697 & 0.829 & 0.892 & 0.884 \\
4\_bullets & 0.613 & 0.661 & 0.795 & 0.867 & 0.850 \\
5\_bullets & 0.600 & 0.647 & 0.784 & 0.853 & 0.829 \\
6\_bullets & 0.573 & 0.627 & 0.756 & 0.829 & 0.807 \\
7\_bullets & 0.555 & 0.606 & 0.741 & 0.819 & 0.791 \\
8\_bullets & 0.543 & 0.596 & 0.724 & 0.804 & 0.769 \\
\bottomrule
\end{tabular}
\end{table}

\begin{table}[htbp]
\centering
\caption{AUC scores for Peer Reviews generated with Qwen3.5 using the N-grams method (n=2).}
\label{tab:pr_random_spans_2}
%\resizebox{\textwidth}{!}{%
\begin{tabular}{l ccccc}
\toprule
Group (vs human) & r=0.1 & r=0.2 & r=0.3 & r=0.4 & r=0.5 \\
\midrule
1\_bullets & 0.898 & 0.966 & 0.976 & 0.982 & 0.984 \\
2\_bullets & 0.874 & 0.939 & 0.951 & 0.967 & 0.968 \\
3\_bullets & 0.842 & 0.916 & 0.930 & 0.948 & 0.952 \\
4\_bullets & 0.815 & 0.875 & 0.900 & 0.924 & 0.928 \\
5\_bullets & 0.793 & 0.860 & 0.891 & 0.909 & 0.912 \\
6\_bullets & 0.805 & 0.853 & 0.874 & 0.905 & 0.904 \\
7\_bullets & 0.776 & 0.825 & 0.861 & 0.883 & 0.888 \\
8\_bullets & 0.771 & 0.833 & 0.847 & 0.875 & 0.880 \\
\bottomrule
\end{tabular}%
%}
\end{table}

\begin{table}[htbp]
\centering
\caption{AUC scores for Peer Reviews generated with Qwen3.5 using the N-grams method (n=3).}
\label{tab:pr_random_spans_3}
%\resizebox{\textwidth}{!}{%
\begin{tabular}{l ccccc}
\toprule
Group (vs human) & r=0.1 & r=0.2 & r=0.3 & r=0.4 & r=0.5 \\
\midrule
1\_bullets & 0.929 & 0.967 & 0.983 & 0.984 & 0.986 \\
2\_bullets & 0.907 & 0.947 & 0.966 & 0.970 & 0.975 \\
3\_bullets & 0.875 & 0.923 & 0.951 & 0.957 & 0.961 \\
4\_bullets & 0.847 & 0.893 & 0.925 & 0.929 & 0.939 \\
5\_bullets & 0.839 & 0.881 & 0.914 & 0.926 & 0.925 \\
6\_bullets & 0.833 & 0.882 & 0.907 & 0.913 & 0.918 \\
7\_bullets & 0.819 & 0.853 & 0.893 & 0.893 & 0.901 \\
8\_bullets & 0.813 & 0.848 & 0.883 & 0.883 & 0.895 \\
\bottomrule
\end{tabular}%
%}
\end{table}

\begin{table}[htbp]
\centering
\caption{AUC scores for Peer Reviews generated with Qwen3.5 using the Surprisal Spans method (Span of length 1).}
\label{tab:pr_surprisal_spans_1}
%\resizebox{\textwidth}{!}{%
\begin{tabular}{l ccccc}
\toprule
Group (vs human) & r=0.1 & r=0.2 & r=0.3 & r=0.4 & r=0.5 \\
\midrule
1\_bullets & 0.674 & 0.827 & 0.923 & 0.965 & 0.977 \\
2\_bullets & 0.664 & 0.778 & 0.881 & 0.933 & 0.951 \\
3\_bullets & 0.682 & 0.783 & 0.862 & 0.902 & 0.923 \\
4\_bullets & 0.652 & 0.738 & 0.809 & 0.869 & 0.890 \\
5\_bullets & 0.641 & 0.716 & 0.784 & 0.839 & 0.864 \\
6\_bullets & 0.602 & 0.681 & 0.765 & 0.825 & 0.848 \\
7\_bullets & 0.601 & 0.661 & 0.744 & 0.804 & 0.829 \\
8\_bullets & 0.580 & 0.653 & 0.719 & 0.784 & 0.815 \\
\bottomrule
\end{tabular}%
%}
\end{table}

\begin{table}[htbp]
\centering
\caption{AUC scores for Peer Reviews generated with Qwen3.5 using the Surprisal Spans method (Span of length 2).}
\label{tab:pr_surprisal_spans_2}
%\resizebox{\textwidth}{!}{%
\begin{tabular}{l ccccc}
\toprule
Group (vs human) & r=0.1 & r=0.2 & r=0.3 & r=0.4 & r=0.5 \\
\midrule
1\_bullets & 0.755 & 0.902 & 0.957 & 0.975 & 0.983 \\
2\_bullets & 0.736 & 0.870 & 0.929 & 0.955 & 0.966 \\
3\_bullets & 0.742 & 0.856 & 0.910 & 0.934 & 0.946 \\
4\_bullets & 0.665 & 0.806 & 0.870 & 0.903 & 0.919 \\
5\_bullets & 0.654 & 0.788 & 0.853 & 0.883 & 0.897 \\
6\_bullets & 0.635 & 0.761 & 0.838 & 0.866 & 0.885 \\
7\_bullets & 0.620 & 0.734 & 0.809 & 0.849 & 0.870 \\
8\_bullets & 0.593 & 0.730 & 0.802 & 0.840 & 0.860 \\
\bottomrule
\end{tabular}%
%}
\end{table}

\begin{table}[htbp]
\centering
\caption{AUC scores for Peer Reviews generated with Qwen3.5 using the Surprisal Spans method (Span of length 3).}
\label{tab:pr_surprisal_spans_3}
%\resizebox{\textwidth}{!}{%
\begin{tabular}{l ccccc}
\toprule
Group (vs human) & r=0.1 & r=0.2 & r=0.3 & r=0.4 & r=0.5 \\
\midrule
1\_bullets & 0.768 & 0.906 & 0.961 & 0.978 & 0.983 \\
2\_bullets & 0.750 & 0.883 & 0.937 & 0.955 & 0.966 \\
3\_bullets & 0.766 & 0.875 & 0.917 & 0.941 & 0.949 \\
4\_bullets & 0.685 & 0.812 & 0.879 & 0.910 & 0.921 \\
5\_bullets & 0.671 & 0.799 & 0.868 & 0.890 & 0.904 \\
6\_bullets & 0.663 & 0.787 & 0.854 & 0.883 & 0.896 \\
7\_bullets & 0.641 & 0.754 & 0.827 & 0.863 & 0.877 \\
8\_bullets & 0.625 & 0.748 & 0.824 & 0.858 & 0.875 \\
\bottomrule
\end{tabular}%
%}
\end{table}

\begin{table}[htbp]
\centering
\caption{AUC scores for Peer Reviews generated with Qwen3.5 using the Top-ranked sentences method.}
\label{tab:pr_extractive}
\begin{tabular}{l ccc}
\toprule
Group (vs human) & 1 Sentence & 3 Sentences & 5 Sentences \\
\midrule
1\_bullets & 0.511 & 0.573 & 0.597 \\
2\_bullets & 0.506 & 0.559 & 0.573 \\
3\_bullets & 0.512 & 0.519 & 0.537 \\
4\_bullets & 0.513 & 0.518 & 0.535 \\
5\_bullets & 0.511 & 0.513 & 0.529 \\
6\_bullets & 0.505 & 0.512 & 0.529 \\
7\_bullets & 0.504 & 0.511 & 0.524 \\
8\_bullets & 0.511 & 0.507 & 0.525 \\
\bottomrule
\end{tabular}
\end{table}

\FloatBarrier
\section{Licenses}

\begin{table*}[htbp]
\centering
\caption{Licenses and terms of use for the external models, datasets, and baseline assets used in this work. For proprietary API models, we report the governing service terms rather than an open-weight model license.}
\footnotesize
\setlength{\tabcolsep}{4pt}
\resizebox{\textwidth}{!}{%
\begin{tabular}{p{3.2cm} p{2.5cm} p{4.2cm} }
\toprule
\textbf{Asset} & \textbf{Type} & \textbf{License / Terms}  \\
\midrule

Llama-3.1 family
& Open-weight model family
& Llama 3.1 Community License
\\

Qwen3 family
& Open-weight model family
& Apache License 2.0
\\

Qwen3.5 family
& Open-weight model family
& Apache License 2.0
\\

Gemma 4 family
& Open-weight model family
& Apache License 2.0
\\

Mistral Small family
& Open-weight model family
& Apache License 2.0
\\

GPT-5.4
& Proprietary API model
& OpenAI API / Service Terms
\\

Gemini 3.1 Pro Preview
& Proprietary API model
& Gemini API Additional Terms of Service
\\

IntelLabs AI Peer Review Detection Benchmark
& Dataset
& Intel Limited Internal Research \& Development Use License Agreement
\\

EditLens checkpoint
& Baseline model
& CC BY-NC-SA 4.0
\\

Binoculars
& Baseline method repository
& BSD 3-Clause License
\\

Fast-DetectGPT
& Baseline method repository
& MIT License
\\

ModernBERT Detector
& Baseline model checkpoint
& Apache License 2.0
\\

\bottomrule
\end{tabular}
}

\label{tab:asset_licenses}
\end{table*}

%%%%%%%%%%%%%%%%%%%%%%%%%%%%%%%%%%%%%%%%%%%%%%%%%%%%%%%%%%%%

\end{document}